\documentclass{article}
\usepackage{arxiv}
\usepackage{amsmath}
\usepackage{amssymb}
\usepackage[utf8]{inputenc} 
\usepackage[T1]{fontenc}    
\usepackage{url}            
\usepackage{booktabs}       
\usepackage{amsfonts}       
\usepackage{nicefrac}       
\usepackage{microtype}      
\usepackage{lipsum}		
\usepackage{graphicx}
\usepackage{subcaption}
\usepackage{makecell}

\usepackage[pagebackref=true,breaklinks=true,letterpaper=true,colorlinks,bookmarks=false]{hyperref}

\title{CLADE: Cycle Loss Augmented Degradation Enhancement for Unpaired Super-Resolution of Anisotropic Medical Images}

\author{
    {\hspace{1mm}Michele Pascale}\\
	Department of Medical Physics\\
	University College London\\
	London, UK \\
	\AND
	\hspace{1mm}Vivek Muthurangu \\
	Institute of Cardiovascular Science\\
	University College London\\
	London, UK \\
 	\And
	\hspace{1mm}Javier Montalt Tordera \\
	Institute of Cardiovascular Science\\
	University College London\\
	London, UK \\
 	\AND
	\hspace{1mm}Heather E Fitzke \\
	Centre for Medical Imaging\\
	University College London\\
	London, UK \\
 	\And
	\hspace{1mm}Gauraang Bhatnagar \\
	Frimley Health NHS Trust\\
	Frimley, UK \\
  	\And
	\hspace{1mm}Stuart A Taylor \\
	UCL NHS Trust\\
	University College London\\
	London, UK \\
  	\AND
	\hspace{1mm}Jennifer Anne Steeden \\
	Institute of Cardiovascular Science\\
	University College London\\
	London, UK \\
}

\date{}

\renewcommand{\shorttitle}{\textit{CLADE: Cycle Loss Augmented Degradation Enhancement}}

\begin{document}
\maketitle

\begin{abstract}
Three-dimensional (3D) imaging is popular in medical applications, however, anisotropic 3D volumes with thick, low-spatial-resolution slices are often acquired to reduce scan times. Deep learning (DL) offers a solution to recover high-resolution features through super-resolution reconstruction (SRR). Unfortunately, paired training data is unavailable in many 3D medical applications and therefore we propose a novel unpaired approach; CLADE (Cycle Loss Augmented Degradation Enhancement). CLADE uses a modified CycleGAN architecture with a cycle-consistent gradient mapping loss, to learn SRR of the low-resolution dimension, from disjoint patches of the high-resolution plane within the anisotropic 3D volume data itself. We show the feasibility of CLADE in abdominal MRI and abdominal CT and demonstrate significant improvements in CLADE image quality over low-resolution volumes and state-of-the-art self-supervised SRR; SMORE (Synthetic Multi-Orientation Resolution Enhancement). Quantitative PIQUE (qualitative perception-based image quality evaluator) scores and quantitative edge sharpness (ES - calculated as the maximum gradient of pixel intensities over a border of interest), showed superior performance for CLADE in both MRI and CT. Qualitatively CLADE had the best overall image quality and highest perceptual ES over the low-resolution volumes and SMORE. This paper demonstrates the potential of using CLADE for super-resolution reconstruction of anisotropic 3D medical imaging data without the need for paired 3D training data.
\end{abstract}


\section{Introduction}
Three-dimensional (3D) Computed Tomography (CT) and Magnetic Resonance Imaging (MRI) have become a vital part of medical diagnosis, enabling comprehensive assessment of anatomy. However, 3D acquisitions are often time-consuming, resulting in long overall scan times and / or artefacts due to physiological or voluntary motion, e.g. breathing. To address these issues, anisotropic 3D imaging has emerged as a common solution, offering faster acquisition times and reduced radiation dose in the case of CT. Although, the resulting volumes have high in-plane spatial resolution, they suffer from poorer spatial resolution in the other planes. Traditional interpolation methods, such as nearest neighbour, bilinear, bicubic, B-spline, or Lanczos resampling, are often employed for visualization. However, these approaches often result in “blocky” edges or blurring, particularly when performing multi-planar reformatting (MPR). An alternative approach is super-resolution reconstruction (SRR), which can effectively recover high-resolution features from low-spatial-resolution image data \cite{1} with high signal-to-noise ratio (SNR) \cite{2}. Whilst conventional SRR methods face challenges such as extended reconstruction times and suboptimal feature recovery, recent advancements in Deep Learning (DL) have transformed SRR in natural images, generating realistic high-resolution images more efficiently than traditional SRR approaches \cite{3,4}.

\section{Related Works}

\noindent\textbf{Supervised DL-SRR} techniques have shown promise in the context of medical imaging, with most methods relying on supervised learning, which requires paired low-resolution (LR) and high-resolution (HR) training data. Robust super resolution has been achieved using supervised super-resolution Convolutional Neural Networks (SR-CNN) for both 3D MRI and CT, with specific implementations including SR-CNN3D \cite{5}, 3D DCSRN (Densely Connected Super-Resolution Networks) \cite{6} and ResCNN (residual convolutional neural network) with increased network depth and residual layers \cite{7}. These have been further extended to enable arbitrary scale super-resolution (ArSRR) \cite{8} and improved performance has been shown by incorporating data from different planes in cardiac MRI \cite{9}. Unfortunately, for many 3D medical applications, paired data is not available preventing the use of supervised DL-SRR approaches.\\ \\
\noindent\textbf{Unsupervised DL-SRR} methods which do not rely on paired training data have been recently introduced for natural images. Many of these rely on Generative Adversarial Networks (GANs), including: CycleGAN \cite{10} and Cycle-in-Cycle GAN \cite{11}. Further advances include: G-GANISR (Gradual generative adversarial network for image super resolution) \cite{12} which uses incremental up-sampling factors with a least square loss function to improve the training stability, SRFeat (Single Image Super-Resolution with Feature Discrimination) \cite{13} which uses an additional discriminator to help generate high-frequency structural features rather than noisy artifacts, and ESRGAN (Enhanced Super-Resolution Generative Adversarial Networks) \cite{14} which uses a deep network with residual-in-residual dense blocks without batch normalization and an enhanced discriminator. Unsupervised SR-GAN architectures have also shown promise in medical imaging. Examples include: EDSSR (Enhanced deep residual networks for single image super-resolution) for anisotropic 3D data \cite{15}, GAN-CIRCLE (GAN Constrained by the Identical, Residual, and Cycle Learning Ensemble) applied to CT images, and UASSR (Unsupervised Arbitrary Scale Super-Resolution Reconstruction) \cite{16} which employs disentangled representation learning to enable arbitrary super-resolution factors. 

However, many unsupervised SRR studies applied to medical images must simulate the low-resolution training data through downsampling of high-resolution images. Unfortunately, this is known to introduce a degradation shift compared to authentic low-resolution images, compromising the resultant SRR quality on prospective data. Some work has been carried out alleviate this problem by also learning the degradation kernel from unpaired LR and HR images; UDEAN (Unsupervised Degradation Adaptation Network) \cite{17}, and a SRR network then learns the mapping from the down-sampled HR images to the original ones. However, in this approach any errors in estimating the degradation kernel might propagate into the final reconstruction.

\section{Contributions}

In this paper, we propose CLADE (Cycle Loss Augmented Degradation Enhancement), an entirely unsupervised DL-SRR approach for anisotropic 3D medical images that learns directly from the anisotropic volumes themselves. This avoids the need to learn the degradation kernel, simplifying the overall DL-SRR scheme. The framework relies on the observation that small two-dimensional (2D) patches extracted from a 3D volume contain similar visual features, irrespective of their orientation. Therefore, we propose a novel 2D patch-based approach that learns HR features from patches in the HR plane of the anisotropic volume, to guide DL-SRR in the LR direction in an unpaired fashion. \\

\newpage

\noindent\textbf{CLADE is based on a Cycle-Consistent Adversarial Network (CycleGAN), with the following modifications:} 

\begin{itemize}
    \item A weight demodulation process in the generator to reduce artefacts when performing patch reconstruction and to stabilise network training.
    \item The addition of a cycle-consistent gradient mapping loss to enforce local edge enhancement within the images. 
\end{itemize}

\noindent\textbf{The specific aims of this paper are:}

\begin{enumerate}
    \item To develop CLADE to perform unsupervised DL-SRR in two challenging exemplar applications: anisotropic 3D MRI and 3D CT images of the abdomen.
    \item To assess the impact of the proposed architectural changes and modified cycle-consistent gradient mapping loss in CLADE.
    \item To assess the resultant image quality from CLADE against a state-of-the-art DL-SRR network, SMORE \cite{18}.
    \item To assess the overall image enhancement achieved by CLADE through quantitative assessment (SNR and edge sharpness) and qualitative image ranking. 
\end{enumerate} 

\section{Methodology}
\subsection{Conventional CycleGAN Architecture}
A conventional Generative Adversarial Network (GAN) \cite{19} consists of a generator network, $G$, which learns to produce an output, $\hat{y}$, from a random input noise tensor, $z \in Z$ (where $Z$ is the noise distribution); $\hat{y} = G(z) \in Y$ (where $Y$ is the target distribution, with elements $y \in Y$). The generator network is paired with a discriminator network, $D$, which learns to distinguish $y$ (true data) from $\hat{y}$ (generated data). The generator and discriminator models are trained in a zero-sum, two-player minimax game with an adversarial loss function, $\mathcal{L}_{adv}$ (Equation \ref{eq:adversarial_loss}):

\begin{flalign}
\min_{G} \max_{D} \mathcal{L}_{\text{adv}}(D, G) = \mathbb{E}_{y \sim p_{\text{data}}(y)} [\log D(y)]\ + \mathbb{E}_{z \sim p_z(z)} [\log (1 - D(G(z)))]
\label{eq:adversarial_loss}
\end{flalign}

The CycleGAN \cite{10} expands this concept to enable image-to-image style transfer between two disjoint image collections $X$ and $Y$ (with images defined as $x \in X$ and $y \in Y$). The CycleGAN attempts to translate features from domain $X$ to domain $Y$ (and vice-versa) using two generators to simultaneously learn the forward and backward generator mapping functions, $G_X: X \rightarrow Y$ and $G_Y: Y \rightarrow X$, respectively. 

\begin{figure}[h]
  \centering
  \includegraphics[width=1\textwidth]{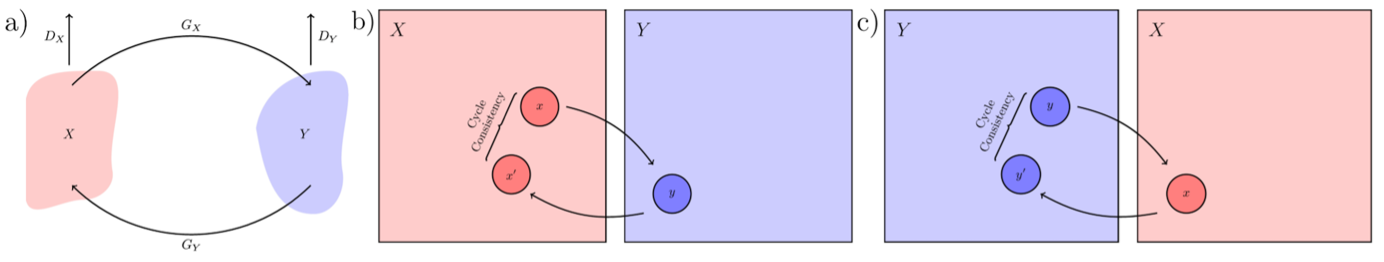}
  \caption{(a) Abstracted CycleGAN model, (b) Forward cycle-consistency loss, (c) Backward cycle-consistency loss. }
  \label{fig:fig1}
\end{figure}

Each generator is paired with a corresponding discriminator, $D_X$ and $D_Y$, respectively, which are trained to distinguish whether an image from the opposite domain is real or generated (see Figure \ref{fig:fig1}(a). The adversarial loss is calculated for the forward and backward generator mappings, as defined in Equation \ref{eq:adversarial_loss_forward} and Equation \ref{eq:adversarial_loss_backward}, respectively:

\begin{flalign}
\mathcal{L}_{adv}(D_Y, G_X) = \mathbb{E}_{y \sim p_{\text{data}}(y)} [\log D_Y(y)]\ + \mathbb{E}_{x \sim p_{\text{data}}(x)} [\log (1 - D_Y(G_X(x)))]
\label{eq:adversarial_loss_forward}
\end{flalign}

\begin{flalign}
\mathcal{L}_{adv}(D_X, G_Y) = \mathbb{E}_{x \sim p_{\text{data}}(x)} [\log D_X(x)]\ + \mathbb{E}_{y \sim p_{\text{data}}(y)} [\log (1 - D_X(G_Y(y)))]
\label{eq:adversarial_loss_backward}
\end{flalign}

Furthermore, to constrain the space of possible mapping functions, the CycleGAN imposes cycle-consistency, which means that the mappings must be approximately bijective and therefore mapping between the two domains should be invertible. \\

Cycle-consistency is achieved through forward and backward cycle-consistency functions, which are defined as $G_X(G_Y(y)) \approx y$ and $G_Y(G_X(x)) \approx x$, respectively (see Figure \ref{fig:fig1}(b) and Figure \ref{fig:fig1}(c)), where the cycle-consistency loss is defined in Equation \ref{eq:cycle_consistency_loss}:

\begin{flalign}
\mathcal{L}_{cyc}(G_X, G_Y) = \mathbb{E}_{x \sim p_{data}(x)} \left[\|G_Y(G_X(x)) - x\|_1\right] + \mathbb{E}_{y \sim p_{data}(y)} \left[\|G_X(G_Y(y)) - y\|_1\right]
\label{eq:cycle_consistency_loss}
\end{flalign}

where $\|\cdot\|_1$ denotes an $L_1$ norm. Therefore, the overall loss function for a conventional CycleGAN, $\mathcal{L}_{GAN}$, consists of the weighted sum of the adversarial losses and the cycle-consistency loss, as defined in Equation \ref{eq:overall_loss}:

\begin{flalign}
\mathcal{L}_{GAN}(D_X, D_Y, G_X, G_Y) = \mathcal{L}_{adv}(D_Y, G_X)\ + 
\mathcal{L}_{adv}(D_X, G_Y)\ + \lambda_{cyc} \mathcal{L}_{cyc}(G_X, G_Y)
\label{eq:overall_loss}
\end{flalign}

where $\lambda_{cyc}$ is the weighting for the cycle-consistency loss term.

\subsection{CLADE Architecture}
\begin{figure}[h]
  \centering
  \includegraphics[width=1\textwidth]{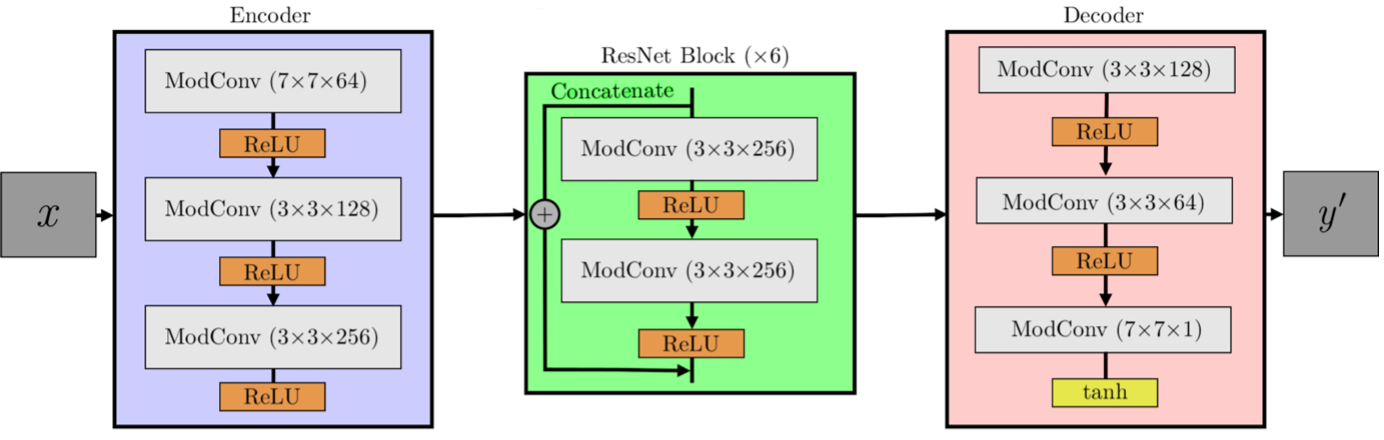}
  \caption{CLADE generator architecture shown for $G_X: X \rightarrow Y$. \texttt{ModConv} denotes the 2D modified convolution layers. Each modified convolutional layer performs the weight-demodulation process.}
  \label{fig:fig2}
\end{figure}

The architecture of CLADE is similar to that of the conventional CycleGAN architecture by Zhu et al.~\cite{10}. The generators ($G_X$ and $G_Y$) each contain an encoder block, six residual blocks, and a decoder block (Figure \ref{fig:fig2}).

However, unlike the conventional CycleGAN, we replaced instance normalisation layers with a weight demodulation process applied directly to the convolutional weights; this aims to reduce the presence of droplet or block-like noise artifacts~\cite{20}. This is similar to that used in StyleGAN2~\cite{21}, where instance normalisation is substituted with a combination of a weight modulation and demodulation process. In StyleGAN2, this eliminates the stylistic influence vector, $s$ (intrinsic to the StyleGAN architecture), from the output convolutional feature maps, $w$, by scaling the weights of the convolutional filters such that $w' = s \cdot w$. As CLADE does not use style vectors, we can assume that $s = 1 = w' = w$, therefore removing the need for the modulation step entirely. However, we retain the demodulation step, prior to the convolution operation, by normalising the convolutional weights themselves by the reciprocal square root of the sum of squared weights (Equation \ref{eq:weight_demodulation}):

\begin{equation}
w''_{ijkl} = \frac{w'_{ijkl}}{\sqrt{\sum_{ijkl} (w'_{ijkl})^2 + \epsilon}},
\label{eq:weight_demodulation}
\end{equation}

where $i$ denotes the index of the output channel, $j$ denotes the horizontal position in the kernel, $k$ denotes the vertical position in the kernel, and $l$ denotes the input channel index; furthermore, we chose $\epsilon = 1 \times 10^{-8}$ as our numerical stability constant to avoid numerical division issues~\cite{21}. \\

\begin{figure}[h]
  \centering
  \includegraphics[width=1\textwidth]{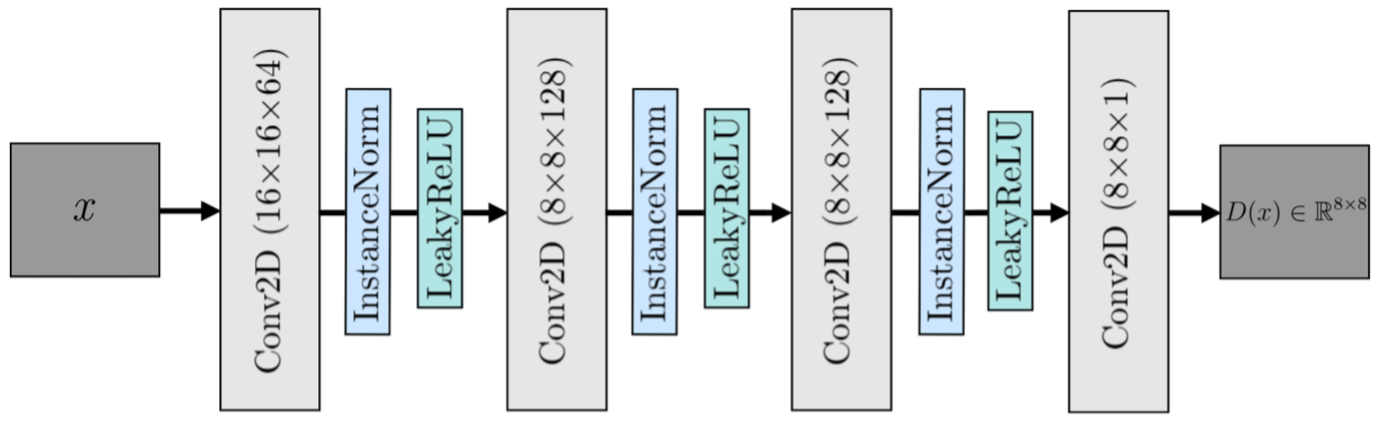}
  \caption{CLADE PatchGAN discriminator architecture.}
  \label{fig:fig3}
\end{figure}

The discriminator networks in CLADE, $D_X$ and $D_Y$, are identical to those in the conventional CycleGAN~\cite{10}, each containing a PatchGAN~\cite{22} (Figure \ref{fig:fig3}).

These discriminators, however, do have instance normalisation layers present, as block-like noise artifacts only occur during image generation and not during the discriminative phase of training.

\subsection{Gradient Mapping Cycle-Consistency Loss}
The conventional CycleGAN architecture has been noted to suffer from deformation errors at the boundaries of soft tissues when employed with anatomical data~\cite{17}. CLADE acts on local patches of the image volume, and therefore the generator has no contextual knowledge of the global image features. \\

Taking advantage of this fact, we opted to investigate the addition of an absolute gradient mapping between the local patches of the cycled high-resolution images to reduce deformation errors, analogous to a Mixed Gradient Loss~\cite{23}. This gradient mapping loss, $\mathcal{L}_{gmap}$, aims to preserve local edge sharpness between gradient maps of input images and ``cycled'' images:

\begin{flalign}
\mathcal{L}_{gmap}(G_X, G_Y)\ = \mathbb{E}_{y \sim p_{data}(y)} \left[\left\| \left(S_x(G_Y(G_X(y))) - S_x(y)\right)+ \left(S_y(G_X(G_Y(y))) - S_y(y\right)\right) \|_{1} \right]],
\label{eq:gradient_mapping_loss}
\end{flalign}

where $S_x$ and $S_y$ denote the image gradient acquired using a Sobel operator in the $x$ and $y$ directions, respectively. The Sobel operator was chosen as it is easy to implement, fast to calculate, and does not require any manual tuning. We combine this gradient mapping loss with the conventional CycleGAN loss, $\mathcal{L}_{GAN}$ (Equation~\ref{eq:overall_loss}), to define the CLADE total loss, $\mathcal{L}_{CLADE}$, as follows:

\begin{flalign}
\mathcal{L}_{CLADE}(D_X, D_Y, G_X, G_Y) = \mathcal{L}_{GAN}(D_X, D_Y, G_X, G_Y) + \lambda_{gmap} \mathcal{L}_{gmap}
\label{eq:clade_total_loss}
\end{flalign}

\subsection{Exemplar Imaging Datasets}
The proposed CLADE approach can be applied to any anisotropic 3D dataset. Therefore, we demonstrated utility on anisotropic 3D MRI images of the abdomen and anisotropic 3D CT images of the abdomen. The SRR networks were trained independently for the CT and MRI datasets due to differences in resolution and visual contrast of the datasets.
\subsubsection{MRI Dataset}
Volumetric Interpolated Breath-hold Examination (VIBE) is a popular MRI technique that enables full abdominal 3D coverage with high in-plane resolution and lower through-plane resolution, within an achievable breath-hold time of $\approx 20$ seconds \cite{24}. VIBE produces T1-weighted anisotropic 3D images and permits simultaneous evaluation of soft tissue and vasculature. \\

In this study, data were retrospectively collected from 60 patients who undertook abdominal 3D VIBE imaging after gadolinium-based contrast agent administration as part of a clinical protocol conducted at University College Hospital, London, UK. The local research ethics committee approved the study (Ref: 10/H0/720/91). Whole abdominal 3D coverage was achieved in a breath-hold of ~21s, with the following imaging parameters: coronal orientation, matrix size $\approx 185 \times 330 \times 23$ (Right-Left, Head-Foot, Anterior-Posterior; RL-HF-AP), field-of-view: $\approx 225 \times 400 \times 150$ mm, resulting in in-plane resolution $\approx1.2 \times 1.2$ mm (range: $0.8$-$1.6$ mm) with thick slices of $\approx 6.6$ mm (range: $5.2$-$7.6$ mm). This resulted in a super-resolution factor of $\approx 5.5$. Of these 60 3D MRI VIBE datasets, 45 were used for training, and 15 were reserved for testing.

\subsubsection{CT Dataset}
The open-source DeepLesion CT dataset \cite{25} comprises of a diverse collection of 32,120 CT slices. In this study, we chose a subset of 60 anisotropic 3D CT volumes of the abdomen, with a comparable anisotropy ratio to the MRI data (i.e. a super-resolution factor of $\approx 5.5$). The data were collected in a transverse orientation with the following imaging parameters: matrix size $512 \times 512$ (RL-AP), $\approx 22$ slices (in HF direction, range: $13$-$83$), with in-plane resolution $\approx 0.95 \times 0.95$ mm (range: $0.90$-$0.98$ mm) and through-plane resolution of $5.0$ mm. This resulted in a super-resolution factor of $\approx 5.3$. The datasets were all cropped to $320 \times 320$ matrix in-plane, to remove air around the body. Of these 60 3D CT datasets, 45 were used for training, and 15 were reserved for testing.

\subsection{Pre-processing and Patching}
Before training the networks, the anisotropic 3D data were pre-processed using cubic-spline interpolation to create isotropic pixels and ensure consistency between the low-resolution and high-resolution data. After pre-processing, each 3D MRI volume contained $\approx128$ (range: $90$-$215$) slices in the low-resolution direction, and each of the CT datasets contained $\approx 118$ slices (range: $67$-$443$) in the low-resolution direction. \\

For training, the interpolated 3D data is separated into 2D slices in both the low-resolution and high-resolution planes. For the MRI data, the coronal plane is high-resolution in both spatial directions, and the sagittal plane was chosen as the low-resolution plane (although the transverse plane could have been chosen). For the CT data, the transverse plane is high-resolution in both spatial directions, and the sagittal plane was chosen as the low-resolution plane (similarly, the coronal plane could have chosen). \\

To create the training dataset, four unpaired, random 2D patches of size ($32 \times 32$) pixels were extracted independently from each low-resolution slice ($x \in X$) and each high-resolution ($y \in Y$) slice of the anisotropic 3D data. The patch size was chosen to be ($32 \times 32$), as this enforces the network to extract local structural information from within the patches and places a restriction on the effective receptive field \cite{26}. This, in turn, reduces the likelihood of learning specific anatomical features and the potential risk for network hallucination. This resulted in a total of 23,068 unpaired patches for training of the MRI networks and 22,520 unpaired patches for training of the CT networks.

\subsection{Network Training and Inference}

All networks were trained in Python3 using Tensorflow 2.0 on an NVIDIA RTX A6000 GPU with 48GB of available memory. Adam \cite{27} was used as the optimizer, with a fixed learning rate of $2 \times 10^{-4}$ and the default momentum coefficient $\beta=0.5$. The weightings for CLADE cycle-consistency ($\lambda_{cyc}$), identity ($\lambda_{ident}$), and gradient-mapping ($\lambda_{gmap}$) losses were optimized by performing hyperparameter optimization using all training data over one epoch. The no-reference image quality metric, PIQUE (Perception-based Image Quality Evaluator) \cite{28}, and subjective image quality over the test dataset were assessed. As PIQUE is a 2D metric, we calculated PIQUE on all 2D slices (in the sagittal orientation) from the volume for each patient in the test dataset, and the mean was used for comparison. Note, a lower PIQUE score represents better image quality. Due to the large amount of training data, the final CLADE networks were trained with 10 epochs, and the models were assessed at each epoch in terms of PIQUE score and subjective image quality. \\

At inference, CLADE can be applied to perform Super-Resolution Reconstruction (SRR) on a full-size anisotropic 3D dataset, which is not of fixed size. Firstly, the anisotropic 3D data must be interpolated in the low-resolution direction (as above) and separated into 2D low-resolution slices. Each slice (of shape $m \times n$) is deconstructed into overlapping ($32 \times 32$) pixel patches and independently passed through the CLADE network. The amount of overlap between consecutive patches is determined by the stride length. We observed a distinct trade-off between image quality and reconstruction time and therefore investigated the effect of stride length on the test datasets, in terms of reconstruction time per volume and image quality using PIQUE scores and subjective image quality. The final images are created using a sliding-window patch reconstruction algorithm that stitches together the resultant patches and then normalizes the images by dividing by the number of times each pixel was subsequently sampled (using the Hadamard product).

\subsection{Comparison of Network Architectures and Losses}
To investigate the impact of the proposed CLADE network modifications and gradient loss on the resultant image quality, we trained the following cycle-consistent unsupervised networks:

\begin{enumerate}
    \item A patch-based Conventional CycleGAN \cite{10} (with six residual blocks)
    \item CLADE architecture with the conventional CycleGAN loss, i.e., $\mathcal{L}_{gmap} = 0$, referred to as “CLADE (no $\mathcal{L}_{gmap}$)”
    \item CLADE architecture with the novel $\mathcal{L}_{CLADE}$ loss, referred to as “CLADE (with $\mathcal{L}_{gmap}$)”
\end{enumerate}

These networks were trained separately for the MRI and CT datasets as described above, with patch-based learning. \\

In addition, we compared the image quality from the conventional CycleGAN, CLADE (no $\mathcal{L}_{gmap}$), and CLADE (with $\mathcal{L}_{gmap}$) to a state-of-the-art DL-SRR approach; SMORE (Synthetic Multi-Orientation Resolution Enhancement)\footnote{The SMORE implementation that was used is publicly accessible here: \url{https://gitlab.com/iacl/smore/-/tree/main}} \cite{29,30}. SMORE is a self-supervised patch-based SRR technique designed to improve the resolution of anisotropic medical images. SMORE does not require any external training data, as it employs a per-volume training and SRR regime, using the high-resolution and low-resolution data that are present in the volumes themselves to both train and super-resolve the image volume. In SMORE, paired training data is simulated through downsampling of the high-resolution plane, and then the trained network is applied to the original low-resolution planes of the volume; SMORE also uses a patch size of ($32 \times 32$).\\

The anisotropic 3D test datasets for MRI and CT were reconstructed using each of the four networks. To assess image consistency across the volumes, the PIQUE scores were calculated from all networks. Although the networks were applied to the low-resolution sagittal plane only, we also calculated PIQUE scores from 2D slices of the volume in the other low-resolution plane (transverse for MRI and coronal for CT).

\subsection{CLADE Image Quality Assessment}
Image quality between the original low-resolution data and resulting images from SMORE, conventional CycleGAN, CLADE (no $\mathcal{L}_{gmap}$), and CLADE (with $\mathcal{L}_{gmap}$) were assessed quantitatively in the test datasets for MRI and CT. \\

Quantitative image quality was measured by calculating edge sharpness (ES) and estimated SNR. ES was quantified on 3D MPR images, as described previously \cite{31}, by measuring the maximum gradient of the pixel intensities perpendicularly across the border of interest. Pixel intensities were filtered using a Savitzky–Golay filter to remove the effect of noise (window width $= 5$ pixels, third-order polynomial) before differentiation. ES was taken as the maximum gradient of the filtered pixel intensities. In the MRI data, ES was measured across four distinct anatomical regions: the abdominal aorta, liver, lower pole of the kidney, and the spleen, and an average ES value was taken for comparison. In the CT data, ES was measured in one region: across a bony structure. Estimated SNR was calculated by dividing the mean signal intensity in a region of interest (ROI) (drawn in the kidney - an area of contrast uptake - in MRI, and in a bony structure in CT) by the standard deviation of the pixel values within a ROI drawn in an area of no signal (air in the stomach in MRI and air outside the body in CT). Quantitative ES and SNR measurements were made using in-house plugins for the OsiriX DICOM viewing platform \cite{32}. \\

Qualitative image quality was also assessed between the original low-resolution data and resulting images from SMORE, and CLADE (with $\mathcal{L}_{gmap}$). Qualitative image ranking was performed by two clinical MR radiologists (S.T and G.B). The observers were presented with the three volumes displayed next to each other in a random order (original low-resolution data, SMORE, and CLADE (with $\mathcal{L}_{gmap}$)), from one volume in the test dataset at a time. The videos were created from both low-resolution planes and assessed independently. The observers were asked to rank their preferred video (1=best, 2=middle, 3=worst, with the possibility of tied ranking), for: 1) Best overall image quality, and 2) Sharpest anatomical edges. The observers’ rankings were averaged for statistical analysis.

\subsection{Statistical Analysis Methodology}
Statistical analyses were performed using Python (\texttt{statsmodels} 0.13.5). Quantitative metrics (PIQUE, ES, and SNR) were compared using one-way repeated measures analysis of variance (ANOVA) with posthoc testing using Tukey Honest Significant Difference (HSD) multiple comparisons of means to determine group-wise relationships and significant results. To ascertain the significance between qualitative ranking, each question was compared for each model using a Friedman Chi-squared test with Neyman posthoc testing (using Python, \texttt{scipy} 1.11.4).

\section{Results}
\subsection{CLADE Network Optimisation}
From the hyperparameter optimisation study, the final CLADE (with $\mathcal{L}_{gmap}$) loss weightings $\lambda_{cyc}$, $\lambda_{ident}$, and $\lambda_{gmap}$ were chosen for the MRI model as 1, 1, 5, and 1, 1, 10 for the CT model. The quantitative PIQUE scores for all hyperparameter models can be found in Table \ref{tab:supporting_inf_1} and Table \ref{tab:supporting_inf_2} for MRI and CT, respectively. Furthermore, representative images can be seen in Figure \ref{fig:supporting_inf_3} and Figure \ref{fig:supporting_inf_4}. Training for the CLADE (with $\mathcal{L}_{gmap}$) models over 10 epochs took $\approx 36$ hours. For each epoch, the quantitative PIQUE scores can be found in Table \ref{tab:supporting_inf_5} and Table \ref{tab:supporting_inf_6} for MRI and CT, respectively. Furthermore, representative images can be found in Figure \ref{fig:supporting_inf_7} and Figure \ref{fig:supporting_inf_8}. The final CLADE (with $\mathcal{L}_{gmap}$) models were chosen from the ninth epoch for MRI and the tenth epoch for CT. \\

The effect of stride length in the patch reconstruction at inference can be seen in terms of PIQUE scores and reconstruction time in Table \ref{tab:supporting_inf_9} and Table \ref{tab:supporting_inf_10} for MRI and CT, respectively, with representative images in Figure \ref{fig:supporting_inf_11} and Figure \ref{fig:supporting_inf_12}. The optimal balance between image quality and reconstruction time for MRI CLADE (with $\mathcal{L}_{gmap}$) was found with a stride length of 12 pixels. This resulted in $\approx 204$ patches per slice (range: $132$-$300$), taking $103 \pm 38$ seconds per MRI volume at inference. An optimal stride length of 6 pixels was chosen for CT CLADE (with $\mathcal{L}_{gmap}$), resulting in $\approx 581$ patches per slice (range: $294$-$1078$), taking $292 \pm 133$ seconds per CT volume at inference. \\

The conventional CycleGAN and CLADE (no $\mathcal{L}_{gmap}$) networks were trained using the same number of epochs, stride length, and the same optimised loss weightings for $\lambda_{cyc}$ and $\lambda_{ident}$, but with $\lambda_{gmap}$ set to 0, for the MRI and CT, respectively. The conventional CycleGAN and CLADE (no $\mathcal{L}_{gmap}$) networks took approximately the same time as CLADE (with $\mathcal{L}_{gmap}$) for training and at inference.

\subsection{Importance of Weight Demodulation}
\begin{figure}[p]
  \centering
  \includegraphics[width=0.45\textwidth]{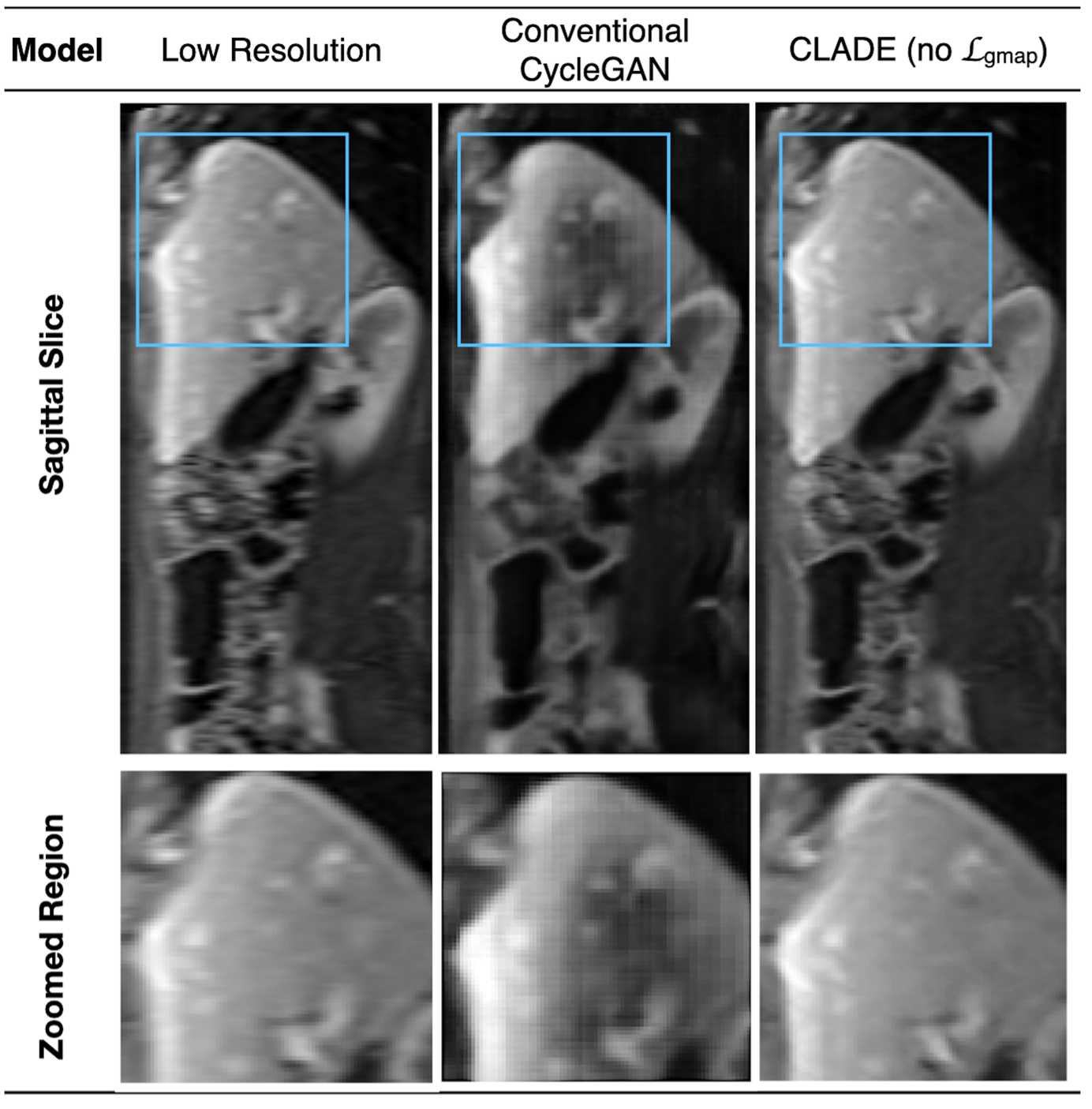}
  \caption{An example low-resolution sagittal MRI slice from one subject in the test data, which shows normalization errors when SRR is applied using a Conventional CycleGAN. These artifacts are removed when SRR is applied using CLADE (no $\mathcal{L}_{gmap}$). Magnified regions within the blue box are displayed beneath each image.}
  \label{fig:fig4}
\end{figure}

The conventional CycleGAN architecture with instance normalization resulted in “blocky” grid artifacts with signal voids in some MRI super-resolved volumes. These errors were successfully removed using the CLADE (no $\mathcal{L}_{gmap}$) as seen in Figure \ref{fig:fig4}, demonstrating the utility of the weight demodulation process in improving MRI image quality. \\

When using the conventional CycleGAN architecture to super-resolve the CT data, we found that the model suffered from mode collapse (see Figure \ref{fig:supporting_inf_13}). This may be due to the lack of visual contrast in the CT datasets, compared to the MRI data, which will likely have made learning these domain mappings a more challenging task.

\subsection{Quantitative Image Quality}
\begin{figure}[p]
  \centering
  \begin{subfigure}{0.46\textwidth}
    \includegraphics[width=\textwidth]{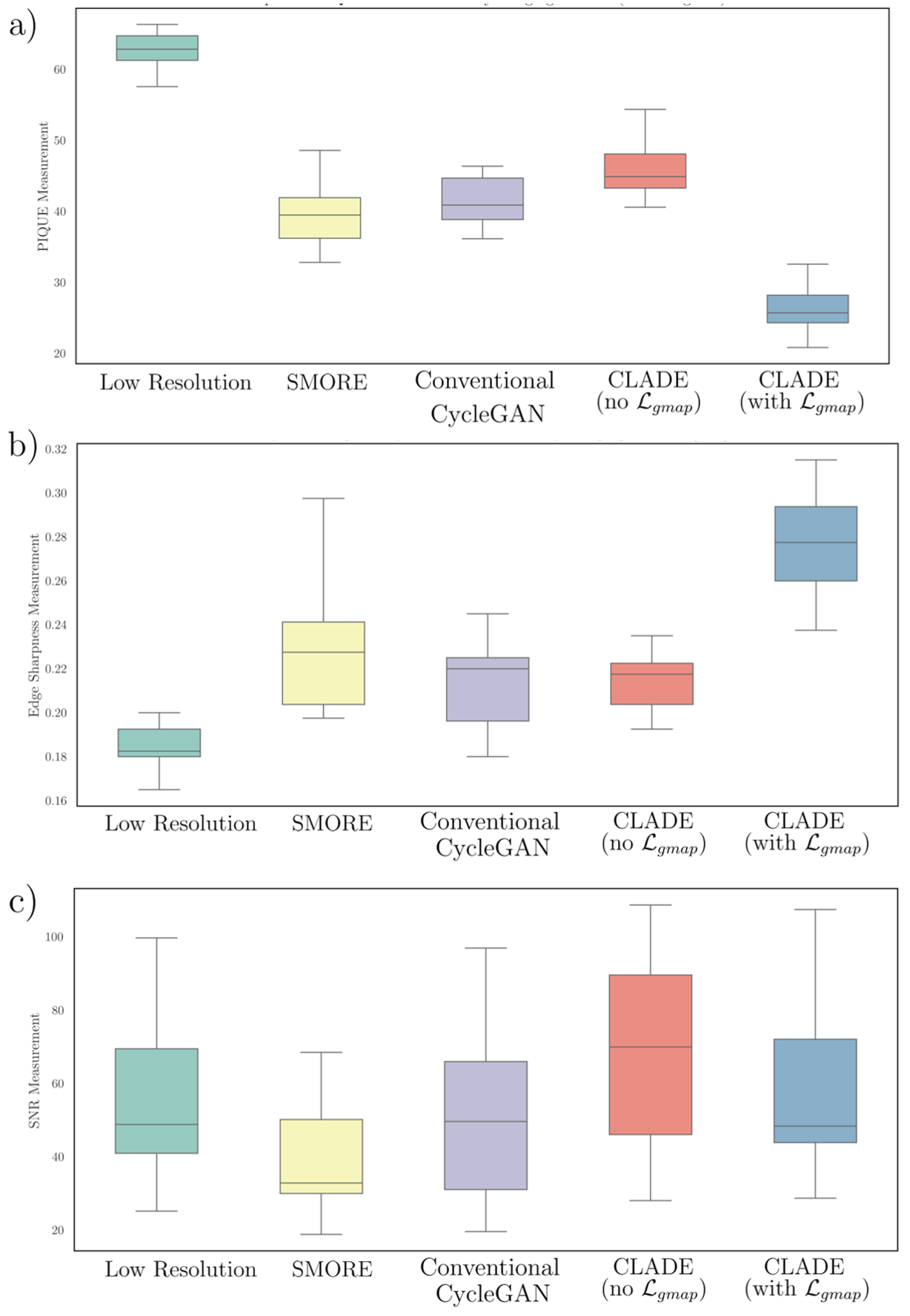}
    \caption{Boxplot for MRI Metrics}
    \label{fig5:sub1}
  \end{subfigure}
  \hfill
  \begin{subfigure}{0.48\textwidth}
    \includegraphics[width=\textwidth]{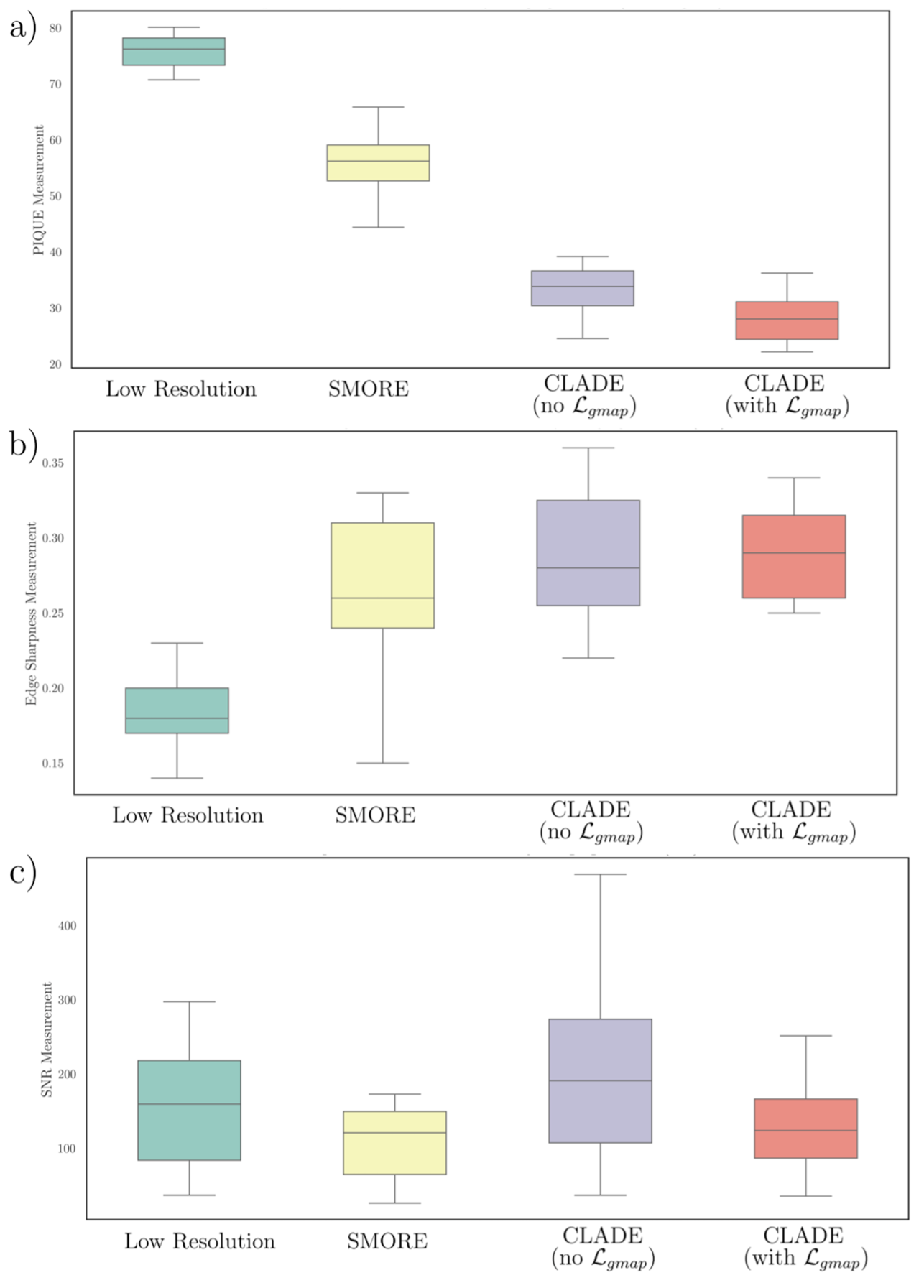}
    \caption{Boxplot for CT Metrics}
    \label{fig5:sub2}
  \end{subfigure}
  \caption{Box Plots for MRI and CT Quantitative Metrics}
  \label{fig:fig5}
\end{figure}

Quantitative PIQUE scores comparing image quality of the original low-resolution volumes with DL-SRR using SMORE, conventional CycleGAN, CLADE (no $\mathcal{L}_{gmap}$) and CLADE ($\mathcal{L}_{gmap}$) can be found in Table \ref{tab:tab1}, and Figure \ref{fig:fig5}(i) and Figure \ref{fig:fig5}(ii) for MRI and CT respectively. Quantitative ES and SNR measurements from all DL-SRR networks are shown in Table \ref{tab:tab2}, and Figure \ref{fig:fig5}(i) and Figure \ref{fig:fig5}(ii), for MRI and CT respectively. \\

The PIQUE scores for all DL-SRR networks were significantly better than the low-resolution volumes for MRI and CT ($p<0.05$), with the exception of the conventional CycleGAN for the CT data that did not train. For both MRI and CT, CLADE ($\mathcal{L}_{gmap}$) produced images with the significantly better PIQUE scores ($p<0.05$) compared to the other DL-SRR methods including SMORE and CLADE ($\mathcal{L}_{gmap} = 0$). Importantly, these trends were the same for PIQUE scores calculated from both low-resolution planes of the MRI and CT volumes.

\newpage 

\begin{table}[htbp]
    \centering
    
    \caption{Quantitative PIQUE scores ($\mu \pm \sigma$) as calculated across all slices in the CT and MRI test volumes. Note that the conventional CycleGAN is not included in the CT results due to model collapse.}
    
    \medskip
    
    \begin{tabular}{lcccc}
        \toprule
        \makecell{Model} & \makecell{\shortstack{MRI PIQUE: \\ Sagittal\\ Orientation**}} & \makecell{\shortstack{MRI PIQUE: \\ Transverse\\ Orientation}} & \makecell{\shortstack{CT PIQUE: \\ Sagittal\\ Orientation**}} & \makecell{\shortstack{CT PIQUE: \\ Coronal\\ Orientation}} \\
        \midrule
        Low-resolution & $62.4 \pm 2.8$ & $64.9 \pm 3.3$ & $74.0 \pm 2.5$ & $75.9 \pm 3.0$ \\
        SMORE & $39.5 \pm 4.5^*$ & $41.0 \pm 4.7^*$ & $52.2 \pm 5.7^*$ & $55.7 \pm 5.5^*$ \\
        Conventional CycleGAN & $41.6 \pm 3.3^*$ & $46.6 \pm 4.4^{*\#}$ & - & - \\
        CLADE (no $\mathcal{L}_{gmap}$) & $45.8 \pm 3.7^{*\#\dagger}$ & $51.1 \pm 3.4^{*\#\dagger}$ & $29.5 \pm 5.5^{*\#}$ & $33.4 \pm 4.3^{*\#}$ \\
        CLADE (with $\mathcal{L}_{gmap}$) & $26.6 \pm 3.8^{*\#\dagger\ddagger}$ & $28.4 \pm 4.0^{*\#\dagger\ddagger}$ & $24.0 \pm 4.9^{*\#\dagger\ddagger}$ & $28.2 \pm 4.5^{*\#\dagger\ddagger}$ \\
        \bottomrule
    \end{tabular}
    
    \medskip
    
    \footnotesize{
        ** Denotes the orientation that the SRR was applied to. \\
        * Denotes statistical differences from low-resolution data ($p < 0.05$). \\
        \# Denotes statistical differences from SMORE ($p < 0.05$). \\
        $^\dagger$ Denotes statistical differences from conventional CycleGAN ($p < 0.05$). \\
        $^\ddagger$ Denotes statistical differences from CLADE (no $\mathcal{L}_{gmap}$) ($p < 0.05$).
    }
    
    \label{tab:tab1}
\end{table}

\begin{table}[htbp]
  \centering
  \caption{Quantitative edge sharpness and estimated SNR ($\mu \pm \sigma$) as calculated across all slices in the CT and MRI test volumes. Note that the conventional CycleGAN is not included in the CT results due to model collapse.}
  \medskip
  \begin{tabular}{lcccccc}
    \toprule
    Model & & Edge Sharpness (mm$^{-1}$) & Signal & Noise & SNR \\
    \midrule
    \multicolumn{6}{l}{\textbf{MRI}} \\
    & Low-resolution & 0.19 $\pm$ 0.04 & 154.5 $\pm$ 34.2 & 3.1 $\pm$ 1.2 & 55.8 $\pm$ 21.5 \\
    & SMORE & 0.23 $\pm$ 0.05$^*$ & 132.5 $\pm$ 46.0 & 3.6 $\pm$ 1.5 & 39.5 $\pm$ 15.4 \\
    & Conventional CycleGAN & 0.21 $\pm$ 0.05$^*$ & 147.6 $\pm$ 46.4 & 3.3 $\pm$ 1.7 & 56.6 $\pm$ 32.7 \\
    & CLADE (no $\mathcal{L}_{gmap}$) & 0.22 $\pm$ 0.04$^*$ & 163.8 $\pm$ 39.8 & 2.6 $\pm$ 1.2 & 73.9 $\pm$ 35.3$^\#$ \\
    & CLADE (with $\mathcal{L}_{gmap}$) & 0.28 $\pm$ 0.04$^{*\#\dagger\ddagger}$ & 162.5 $\pm$ 36.6 & 3.2 $\pm$ 1.3 & 56.7 $\pm$ 21.9 \\
    \midrule
    \multicolumn{6}{l}{\textbf{CT}} \\
    & Low-resolution & 0.19 $\pm$ 0.03 & 167.0 $\pm$ 53.5 & 1.5 $\pm$ 1.2 & 153.6 $\pm$ 81.5 \\
    & SMORE & 0.26 $\pm$ 0.05$^*$ & 101.4 $\pm$ 41.5$^*$ & 1.1 $\pm$ 0.9 & 118.2 $\pm$ 64.4 \\
    & CLADE (no $\mathcal{L}_{gmap}$) & 0.29 $\pm$ 0.04$^*$ & 167.1 $\pm$ 50.4$^\#$ & 1.7 $\pm$ 1.2 & 196.4 $\pm$ 124.3 \\
    & CLADE (with $\mathcal{L}_{gmap}$) & 0.29 $\pm$ 0.03$^*$ & 167.3 $\pm$ 53.8$^\#$ & 1.4 $\pm$ 1.4 & 130.6 $\pm$ 68.4 \\
    \bottomrule
  \end{tabular}
  \label{tab:tab2}
  
  \smallskip
  
  \footnotesize{
  * Denotes statistical differences from low-resolution data ($p < 0.05$). \\
  $\#$ Denotes statistical differences from SMORE ($p < 0.05$). \\
  $^\dagger$ Denotes statistical differences from conventional CycleGAN ($p < 0.05$). \\
  $^\ddagger$ Denotes statistical differences from CLADE (no $\mathcal{L}_{gmap}$) ($p < 0.05$).
  }
\end{table}

The ES was significantly higher for all DL-SRR networks compared to the low-resolution volumes ($p < 0.05$) for both MRI and CT data. In MRI, CLADE ($\mathcal{L}_{gmap}$) was found to have a significantly higher ES than SMORE, the conventional CycleGAN and CLADE (no $\mathcal{L}_{gmap}$). No significant differences were found in the estimated SNR between the low-resolution data and any of the DL-SRR networks in either the MRI or CT data. All networks had similar estimated SNR, with the only significant difference found between MRI CLADE (no $\mathcal{L}_{gmap}$) and MRI SMORE (due to higher signal and lower noise measurements in CLADE (no $\mathcal{L}_{gmap}$).

\subsection{Qualitative Image Quality}
\begin{figure}[p]
  \centering
  \includegraphics[width=0.57\textwidth]{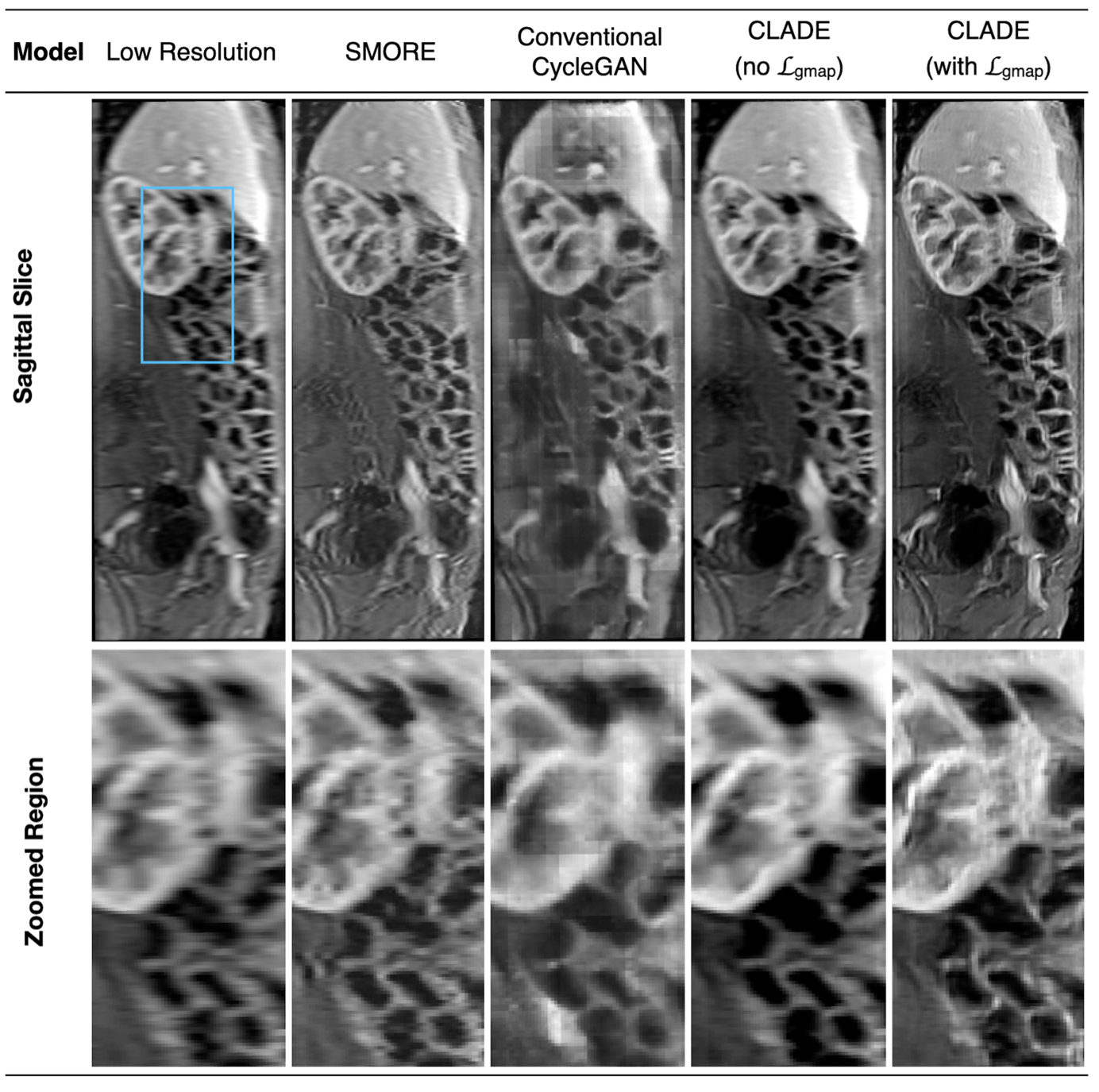}
  \caption{An example of sagittal image quality from SRR applied to a low-resolution MRI data from one subject in the test dataset. Note that the SRR models are applied in the low-resolution sagittal plane for the MRI data. Magnified regions within the blue box are displayed beneath each image. }
  \label{fig:fig6}
\end{figure}

\begin{figure}[h!]
  \centering
  \includegraphics[width=0.57\textwidth]{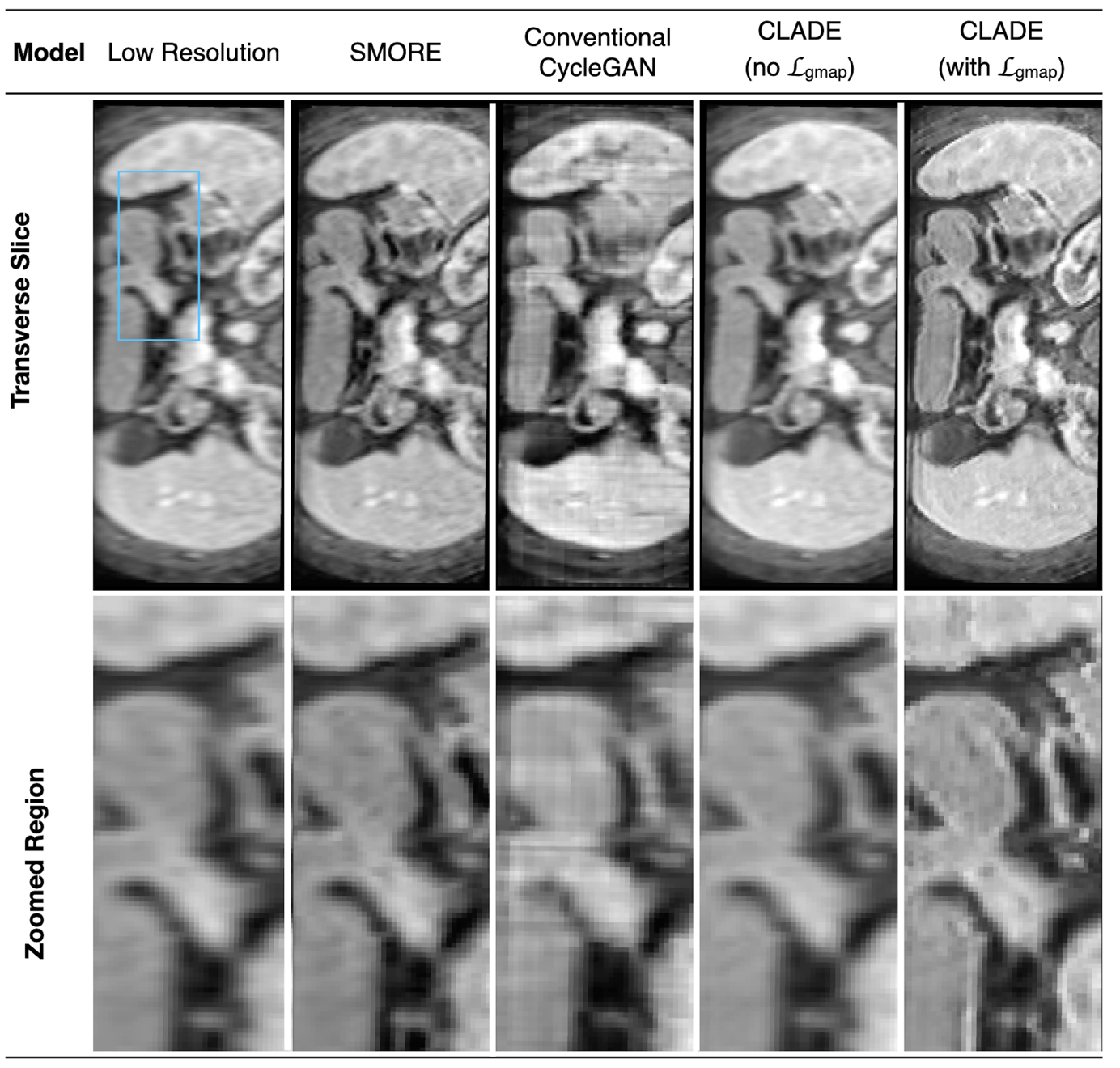}
  \caption{An example of transverse image quality from SRR applied to a low-resolution MRI data from one subject in the test dataset. Note that the SRR models are applied in the low-resolution sagittal plane for the MRI data, where this figure shows the resulting volume reformatted in the (low-resolution) transverse plane. Magnified regions within the blue box are displayed beneath each image. }
  \label{fig:fig7}
\end{figure}

Qualitative image rankings are shown in Table 3, for both MRI and CT. In the MRI data, the qualitative image quality and edge sharpness were ranked significantly better for SMORE and CLADE ($\mathcal{L}_{gmap}$) over the low-resolution volumes, in both orientations. CLADE ($\mathcal{L}_{gmap}$) was found to be significantly better than SMORE in terms of qualitative image quality and edge sharpness in the sagittal plane, but not the transverse plane. These results are reflected in Figure \ref{fig:fig6} and Figure \ref{fig:fig7} which show example image quality from one MRI volume for the DL-SRR networks in both the sagittal and transverse orientations, respectively. \\

Similar to Figure 4, the conventional CycleGAN can be seen to suffer with significant blocky artefacts due to the use of instance normalization, particularly in the sagittal orientation. Importantly, CLADE ($\mathcal{L}_{gmap}$) can be seen to recover more high-resolution features than SMORE, conventional CycleGAN and CLADE (no $\mathcal{L}_{gmap}$) for the MRI data.\\

\begin{figure}[p]
  \centering
  \includegraphics[width=0.4\textwidth]{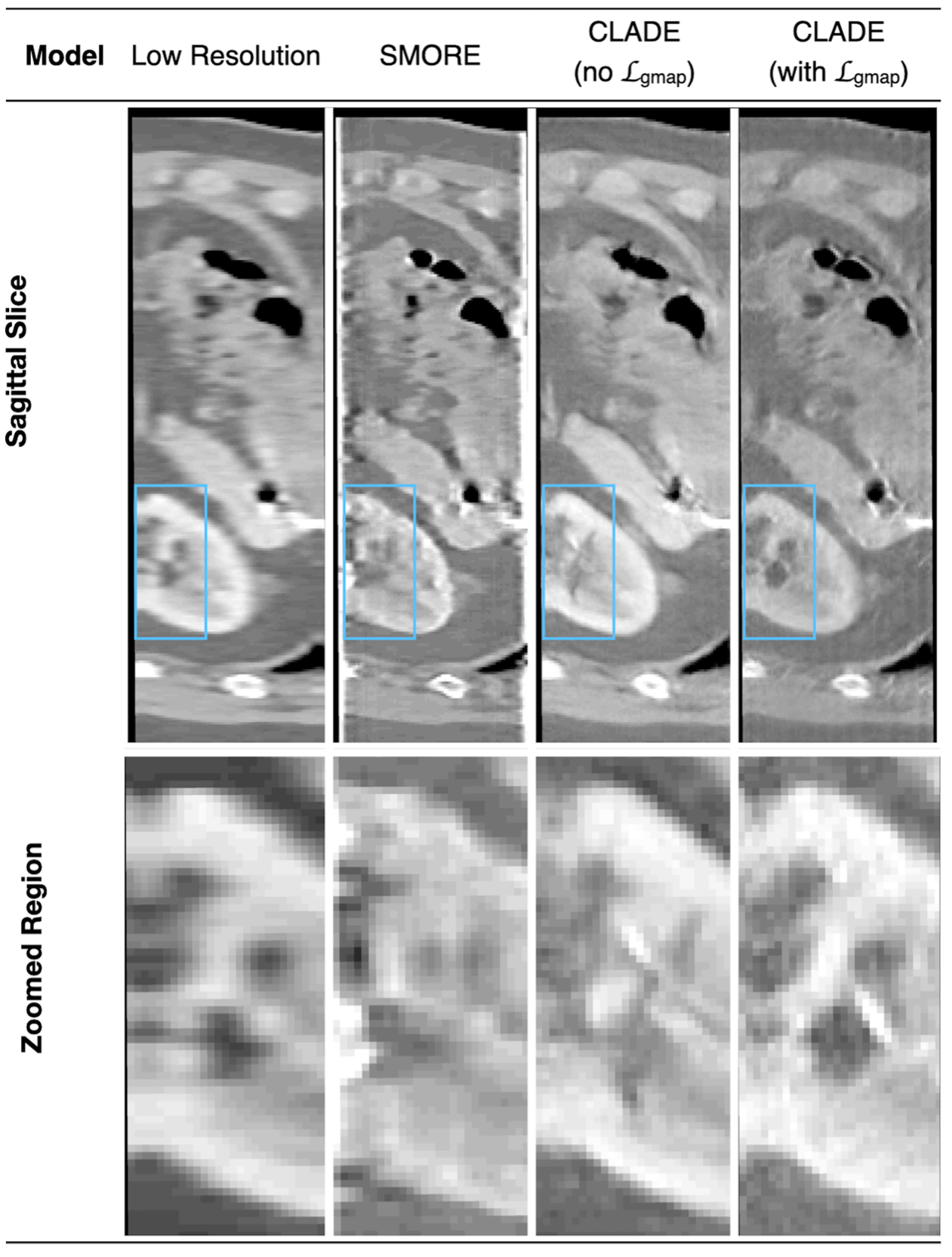}
  \caption{An example of sagittal image quality from SRR applied to a low-resolution CT data from one subject in the test dataset. Note that the SRR models are applied in the low-resolution sagittal plane for the CT data. Magnified regions within the blue box are displayed beneath each image.}
  \label{fig:fig8}
\end{figure}

\begin{figure}[p]
  \centering
  \includegraphics[width=0.4\textwidth]{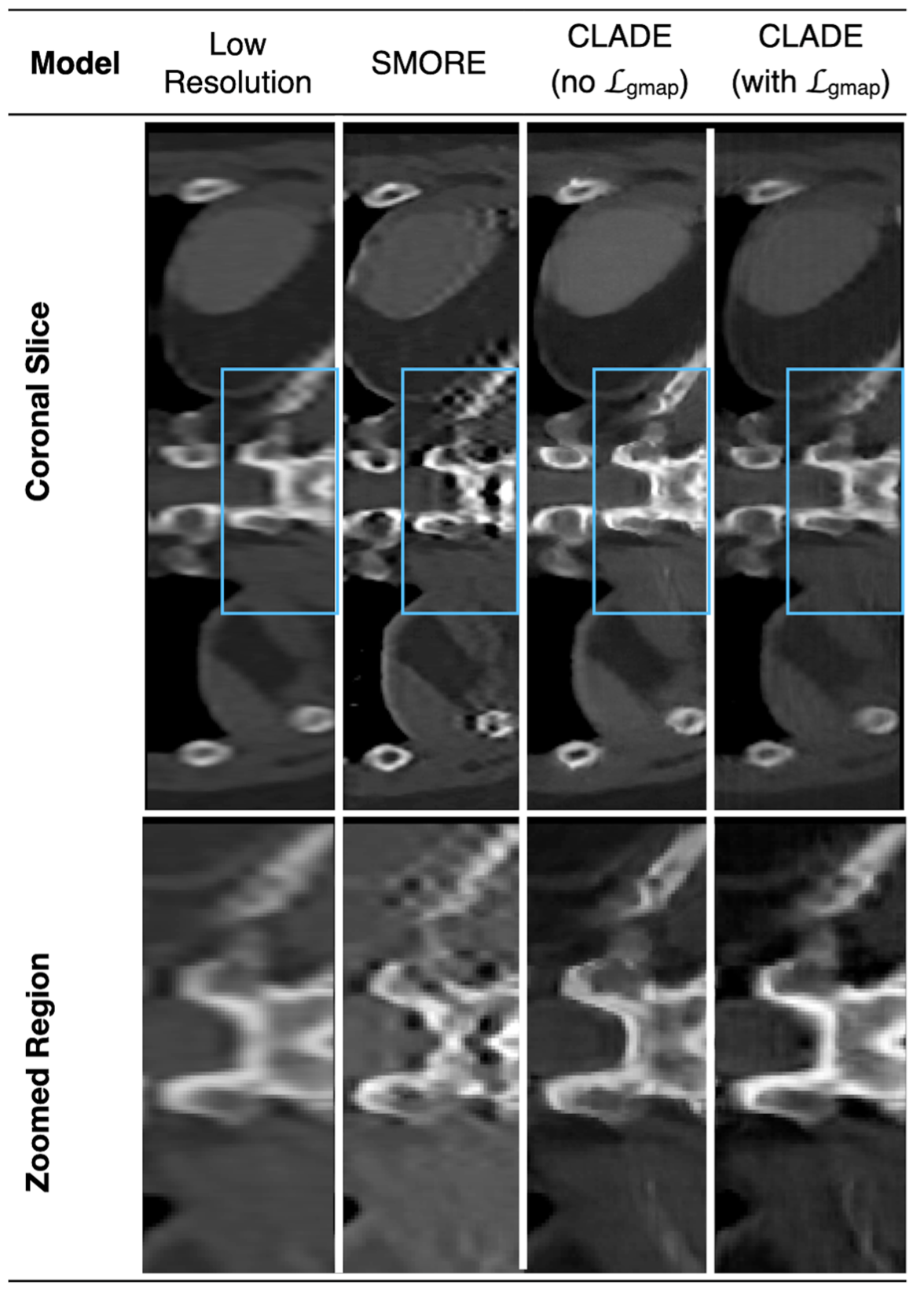}
  \caption{An example of coronal image quality from SRR applied to a low-resolution CT data from one subject in the test dataset. Note that the SRR models are applied in the low-resolution sagittal plane for the CT data, where this figure shows the resulting volume reformatted in the (low-resolution) coronal plane. Magnified regions within the blue box are displayed beneath each image.}
  \label{fig:fig9}
\end{figure}

\begin{figure}[h!]
  \centering
  \includegraphics[width=1\textwidth]{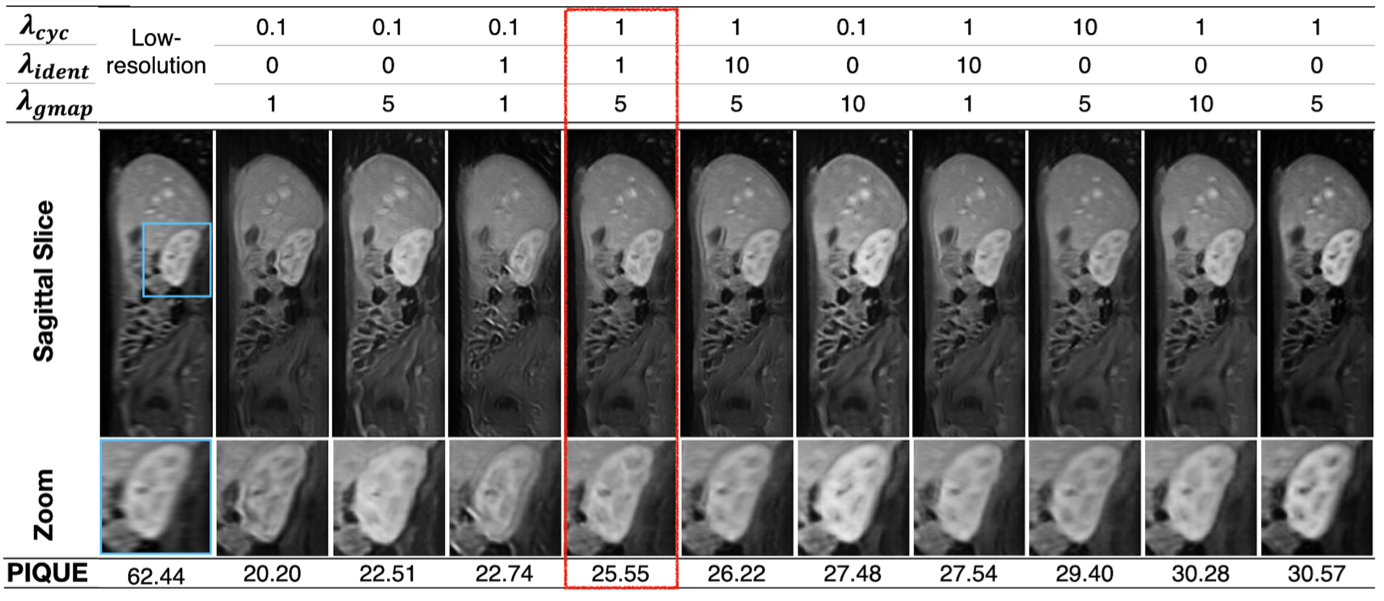}
  \caption{Example image quality from one subject in the MRI test dataset, with SRR applied using CLADE (with $\mathcal{L}_{gmap}$) from the MRI hyperparameter optimisation study, for the 10 models with the lowest PIQUE scores. Magnified regions within the blue box are displayed beneath each image. PIQUE score shows the mean value across all 15 volumes in the MRI test dataset. The red box indicates the chosen final loss weighting based on PIQUE scores (see Table \ref{tab:supporting_inf_1}) as well as visual image quality.}
  \label{fig:supporting_inf_3}
\end{figure}

\begin{figure}[h!]
  \centering
  \includegraphics[width=1\textwidth]{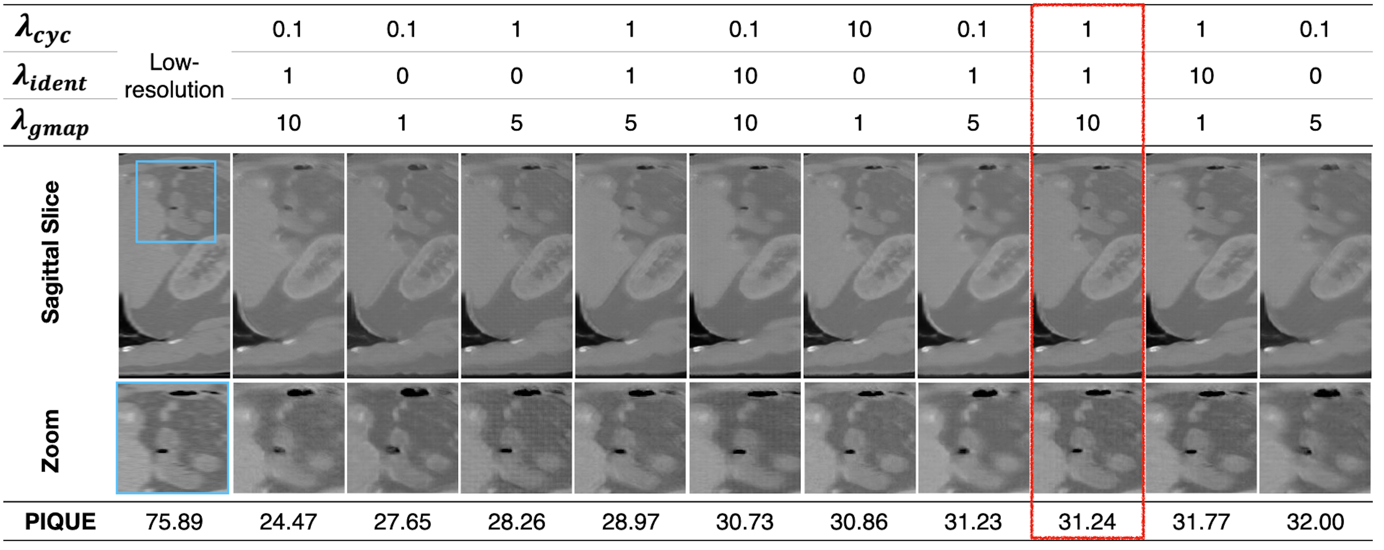}
  \caption{Example image quality from one subject in the CT test dataset, with SRR applied using CLADE (with $\mathcal{L}_{gmap}$) from the CT hyperparameter optimisation study, for the 10 models with the lowest PIQUE scores. Magnified regions within the blue box are displayed beneath each image. PIQUE score shows the mean value across all 15 volumes in the CT test dataset. The red box indicates the chosen final loss weighting based on PIQUE scores (see Table \ref{tab:supporting_inf_2}) as well as visual image quality.}
  \label{fig:supporting_inf_4}
\end{figure}

\begin{figure}[p]
  \centering
  \includegraphics[width=1\textwidth]{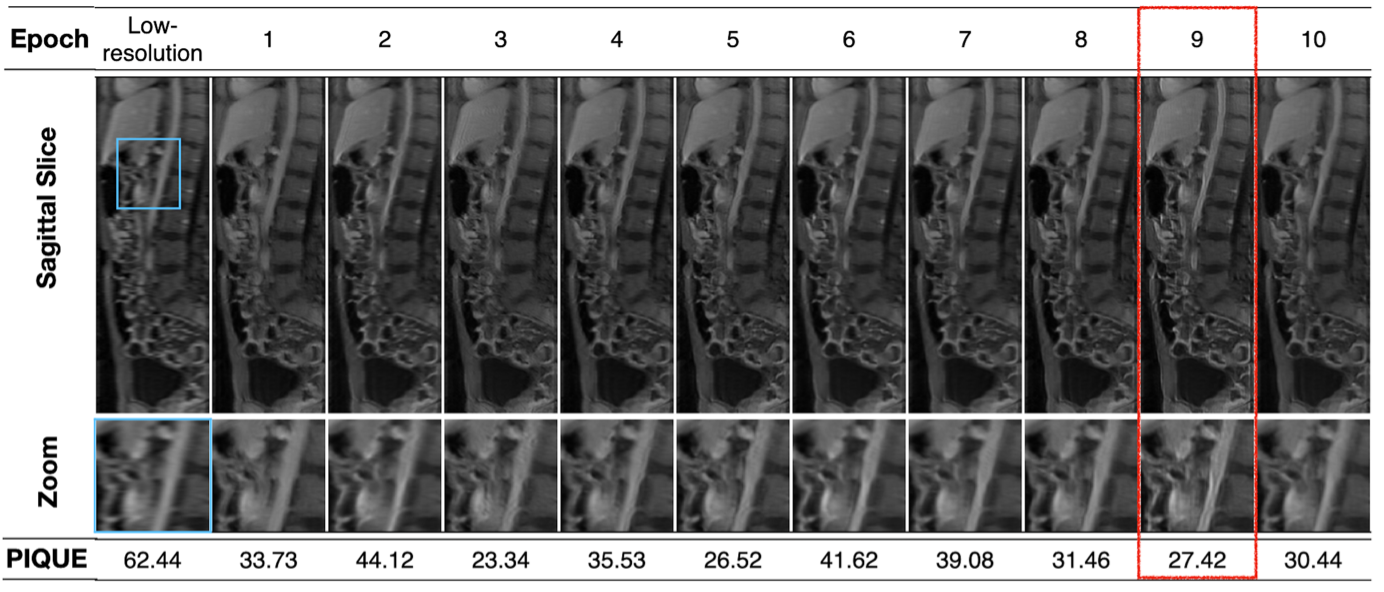}
  \caption{Example image quality from one subject in the MRI test dataset, at each epoch when training the final MRI CLADE (with $\mathcal{L}_{gmap}$) ($\lambda_{cyc} = 1, \lambda_{ident} = 1, \lambda_{cyc} = 5$). Magnified regions within the blue box are displayed beneath each image. PIQUE score shows the mean value across all 15 volumes in the MRI test dataset. The red box indicates the chosen epoch based on PIQUE scores (see Table \ref{tab:supporting_inf_5}) as well as visual image quality.}
  \label{fig:supporting_inf_7}
\end{figure}

\begin{figure}[p]
  \centering
  \includegraphics[width=1\textwidth]{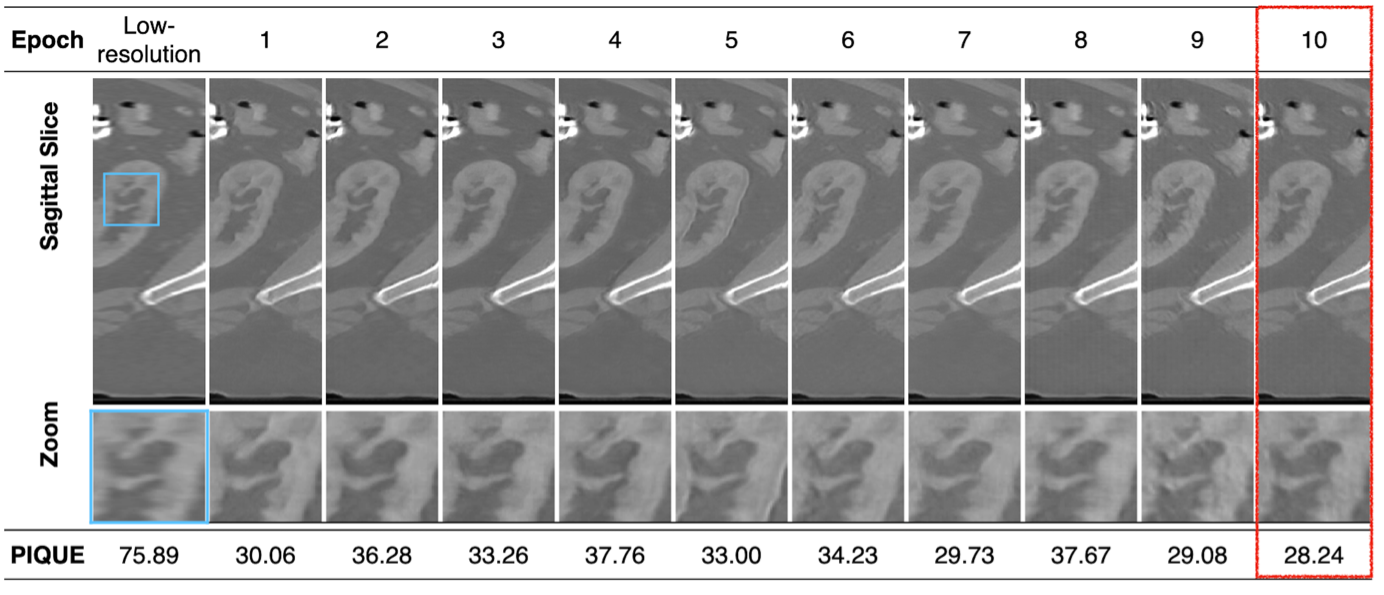}
  \caption{Example image quality from one subject in the CT test dataset, at each epoch when training the final CT CLADE (with $\mathcal{L}_{gmap}$) ($\lambda_{cyc} = 1, \lambda_{ident} = 1, \lambda_{cyc} = 10$). Magnified regions within the blue box are displayed beneath each image. PIQUE score shows the mean value across all 15 volumes in the CT test dataset. The red box indicates the chosen epoch based on PIQUE scores (see Table \ref{tab:supporting_inf_6}) as well as visual image quality.}
  \label{fig:supporting_inf_8}
\end{figure}

\begin{figure}[p]
  \centering
  \includegraphics[width=0.75\textwidth]{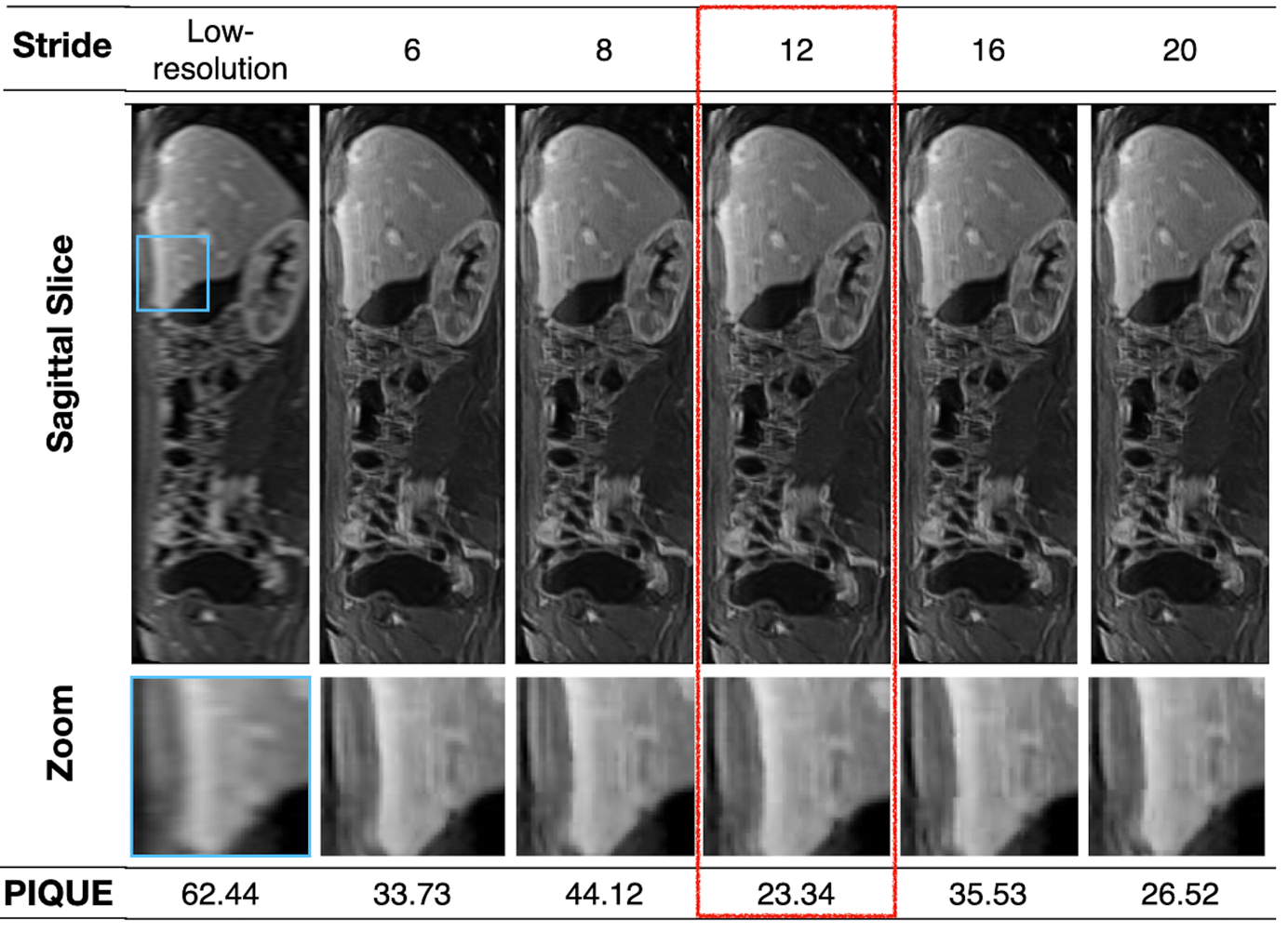}
  \caption{Example image quality from one subject in the MRI test dataset, at each stride length, using the final MRI CLADE (with $\mathcal{L}_{gmap}$) ($\lambda_{cyc} = 1, \lambda_{ident} = 1, \lambda_{cyc} = 5$) during inference. Magnified regions within the blue box are displayed beneath each image. PIQUE score shows the mean value across all 15 volumes in the MRI test dataset. The red box indicates the chosen epoch based on PIQUE score, inference time and visual image quality (see Table \ref{tab:supporting_inf_9}).}
  \label{fig:supporting_inf_11}
\end{figure}

\begin{figure}[p]
  \centering
  \includegraphics[width=0.75\textwidth]{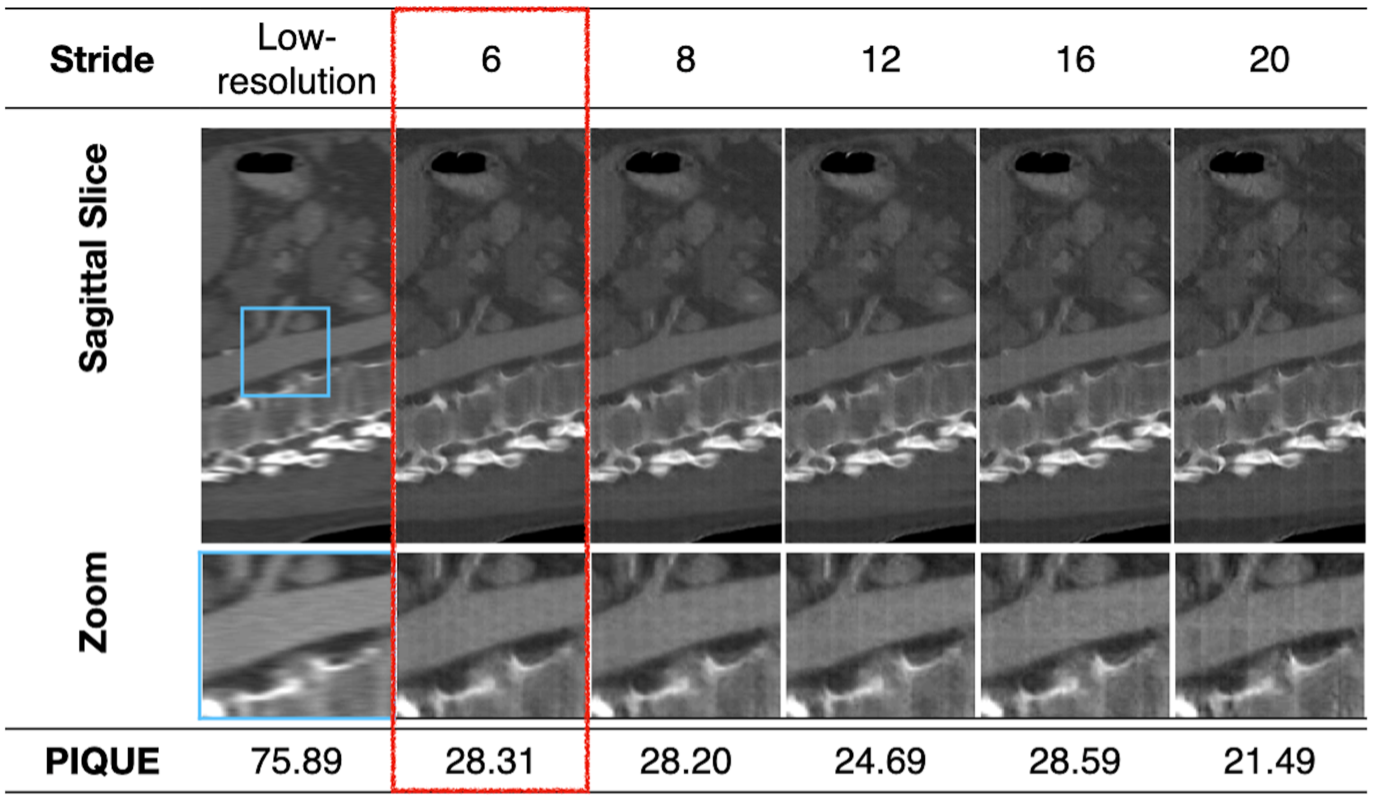}
  \caption{Example image quality from one subject in the CT test dataset, at each stride length, using the final CT CLADE (with $\mathcal{L}_{gmap}$) ($\lambda_{cyc} = 1, \lambda_{ident} = 1, \lambda_{cyc} = 10$) during inference. Magnified regions within the blue box are displayed beneath each image. PIQUE score shows the mean value across all 15 volumes in the CT test dataset. The red box indicates the chosen epoch based on PIQUE score, inference time and visual image quality (see Table \ref{tab:supporting_inf_10}).}
  \label{fig:supporting_inf_12}
\end{figure}

\begin{figure}[p]
  \centering
  \includegraphics[width=0.75\textwidth]{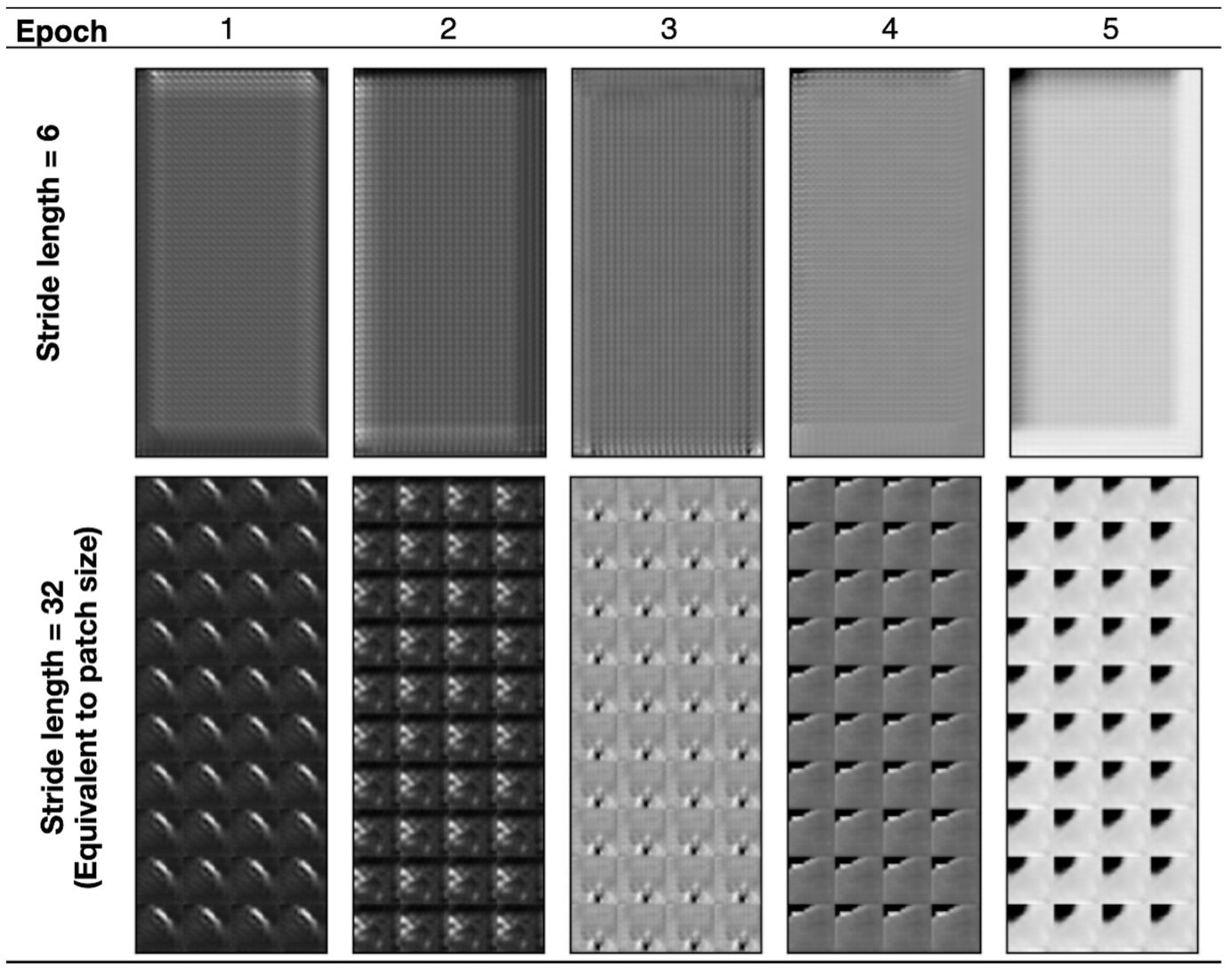}
  \caption{The conventional CycleGAN suffered from mode collapse when trained using the CT data. The top row shows the patched image reconstruction from the model at epochs 1-5, on one of the test CT volumes, with a stride of 6 (as used for CLADE). The bottom row shows that where there are no overlapping patches (i.e. stride length is equal to the patch size), then all patches remain identical, demonstrating mode collapse.}
  \label{fig:supporting_inf_13}
\end{figure}

In the CT data SMORE images were ranked worse than the low-resolution images for qualitative image quality, but better for edge sharpness, in both planes. CT CLADE ($\mathcal{L}_{gmap}$) had significantly better qualitative image quality than both low-resolution volumes and SMORE, in both orientations. CT CLADE ($\mathcal{L}_{gmap}$) was also ranked significantly better for qualitative edge sharpness over the low-resolution images, in both orientations. These results are reflected in Figure \ref{fig:fig8} and Figure \ref{fig:fig9} which show example image quality from one CT volume for the DL-SRR networks in both the sagittal and coronal orientations, respectively. They show that all networks recover some of the high-resolution features in the CT data, however CLADE ($\mathcal{L}_{gmap}$) exhibits fewer artefacts than SMORE and shows less hallucinations than the CLADE (no $\mathcal{L}_{gmap}$) network.

\section{Discussion and Limitations}
In this study, we propose a novel approach, CLADE, to enable super-resolution reconstruction of anisotropic 3D medical images without the need for paired training data. Instead, CLADE relies on the idea that 2D patches in medical images have similar visual features regardless of orientation. CLADE leverages this idea to learn high-resolution features from 2D patches taken from the high-resolution plane, enabling super-resolution of unpaired 2D patches in the low-resolution dimension. We have demonstrated in anisotropic 3D abdominal MRI and CT volumes that CLADE is able to recover high-resolution features in the low-resolution planes. Interestingly, although we applied CLADE to 2D patches taken from just one of the low-resolution planes, the PIQUE scores and qualitative scores were similar when calculated across slices taken from either low-resolution plane of the reconstructed volume, for both MRI and CT. This important finding demonstrates that CLADE improves image quality in all imaging planes without the need to explicitly apply it in both planes. \\

\noindent \textbf{The main findings of this study were:}
\begin{enumerate}
    \item Unsupervised super-resolution reconstruction of anisotropic 3D MRI and CT medical images is feasible using CLADE.
    \item The modifications in the CLADE architecture and losses enabled improved spatial resolution, reduced artefacts, and more stable training compared to a patch-based conventional CycleGAN.
    \item Qualitative image quality and quantitative ranking improved after CLADE super-resolution reconstruction compared to the original anisotropic 3D images.
    \item CLADE outperformed state-of-the-art SMORE in terms of qualitative image quality and quantitative ranking.
\end{enumerate}

The main difference between CLADE and the conventional CycleGAN is the inclusion of both a weight demodulation process and gradient-mapping loss. We demonstrated the efficacy of the weight demodulation process to correct normalisation artifacts that occur in reconstructions using the conventional-CycleGAN approach and to improve stabilisation of the training. Specifically, the conventional-CycleGAN suffered from mode collapse for the CT data, while MRI data exhibited blocky artifacts and poorer visual quality compared to CLADE. A gradient-mapping loss was also leveraged in CLADE ($\mathcal{L}_{gmap}$) to promote generation of images with well-defined edges. We showed that CLADE ($\mathcal{L}_{gmap}$) resulted in significant improvements in quantitative PIQUE scores, quantitative edge sharpness, qualitative overall image quality, and qualitative edge sharpness over the original low-resolution images. Importantly, CLADE preserved SNR in the images and did not cause any additional artifacts.

We also compared CLADE to SMORE, which is a state-of-the-art DL-SRR network. We found that CLADE ($\mathcal{L}_{gmap}$) significantly outperformed SMORE in terms of quantitative PIQUE scores and qualitative image quality for both MRI and CT, as well as quantitative and qualitative edge sharpness for MRI.

The main limitation of this study was that we were unable to perform any task-specific evaluations (such as performance on registration, detection, and segmentation). This is because high-resolution isotropic data were not acquired, preventing comparison to ‘ground truth’ data. Thus, in the future, it will be necessary to perform prospective studies of CLADE where results can be compared to high-resolution isotropic 3D data.

Another limitation of this study was that CLADE does not explicitly suppress motion artifacts that may occur during abdominal MRI or CT. Thus, any artifacts (e.g., ghosting on MRI) might be accentuated by the super-resolution reconstruction. A further issue was that the number of test datasets was small, and further work is required to assess image quality from more sites, vendors, and imaging contrasts. This is vital to ensure that CLADE is widely applicable, although the fact that it works on both MRI and CT data does suggest that CLADE is generalisable. It should also be noted that CLADE was only trained with patches extracted from only one of the low-resolution dimensions. Future work should explore the effect of training with patches from both of the low-resolution mappings to see if this would improve image quality.

Finally, as CLADE does not use paired training data, it is not possible to calculate structural similarity metrics (SSIM) or other common image quality measures. Instead, we used PIQUE to assess the image quality. Other no-reference perceptual quality metrics exist, such as NIQE (natural image quality evaluator) and BRISQUE (Blind/Referenceless Image Spatial Quality Evaluator); however, these were trained on natural images and are not suitable for medical imaging data.

\section{Conclusions}
We have developed an unsupervised super-resolution reconstruction technique, CLADE, which enables the recovery of high-resolution features from anisotropic 3D medical images. We have demonstrated the technique in abdominal MRI and CT data. This enables improved visualisation of structures in the low-resolution dimension without the need for any changes to current imaging protocols.

\section{Acknowledgement}
This work was supported by UK Research and Innovation (MR/S032290/1); Innovate UK (105860); the EPSRC-funded UCL Centre for Doctoral Training in Intelligent, Integrated Imaging in Healthcare (i4health) (EP/S021930/1); and the Department of Health’s NIHR-funded Biomedical Research Centre at Great Ormond Street Hospital.

\newpage

\newpage

\begin{table}[p]
    \centering
    \caption{PIQUE scores from MRI CLADE (with $\mathcal{L}_{gmap}$) hyperparameter optimisation. The final chosen loss weighting is outlined in bold, based on PIQUE as calculated across all 15 volumes in the MRI test dataset, shown in ascending PIQUE order, as well as visual image quality (see Figure \ref{fig:supporting_inf_3}). Each model is trained for 1 epoch, and images are assessed with a stride length of 6. The loss weightings are chosen from the following parameter sets: $\lambda_{gmap} \in \{0.1, 1, 10\}$, $\lambda_{cyc} \in \{0, 1, 10\}$ and $\lambda_{ident} \in \{1, 5, 10\}$. \\
   }
    \medskip
    \begin{tabular}{cccccc}
        \toprule
        $\lambda_{cyc}$ & $\lambda_{ident}$ & $\lambda_{gmap}$ & PIQUE ($\mu$ $\pm$ $\sigma$) \\
        \midrule
        0.1 & 0 & 1 & $20.20 \pm 2.98$ \\
        0.1 & 0 & 5 & $22.51 \pm 4.28$ \\
        0.1 & 1 & 1 & $22.74 \pm 4.76$ \\
        0.1 & 0 & 10 & $27.48 \pm 3.94$ \\
        0.1 & 1 & 10 & $30.98 \pm 5.34$ \\
        0.1 & 10 & 1 & $31.25 \pm 4.66$ \\
        0.1 & 10 & 5 & $31.66 \pm 4.28$ \\
        0.1 & 1 & 5 & $32.15 \pm 4.17$ \\
        0.1 & 10 & 10 & $35.25 \pm 4.90$ \\
        1 & 0 & 1 & $31.34 \pm 4.37$ \\
        1 & 0 & 5 & $30.57 \pm 5.10$ \\
        1 & 0 & 10 & $30.28 \pm 4.94$ \\
        1 & 1 & 1 & $37.78 \pm 4.51$ \\
        \textbf{1} & \textbf{1} & \textbf{5} & $\mathbf{25.55 \pm 4.71}$ \\
        1 & 1 & 10 & $31.26 \pm 4.66$ \\
        1 & 10 & 1 & $27.54 \pm 4.88$ \\
        1 & 10 & 5 & $26.22 \pm 4.62$ \\
        1 & 10 & 10 & $32.05 \pm 4.89$ \\
        10 & 0 & 1 & $44.94 \pm 6.23$ \\
        10 & 0 & 5 & $29.40 \pm 5.13$ \\
        10 & 0 & 10 & $38.62 \pm 5.70$ \\
        10 & 1 & 1 & $36.23 \pm 5.55$ \\
        10 & 1 & 5 & $34.67 \pm 5.13$ \\
        10 & 1 & 10 & $38.38 \pm 5.83$ \\
        10 & 10 & 1 & $36.31 \pm 5.18$ \\
        10 & 10 & 5 & $36.71 \pm 5.81$ \\
        10 & 10 & 10 & $36.72 \pm 5.45$ \\
        \bottomrule
    \end{tabular}
    \label{tab:supporting_inf_1}
\end{table}

\begin{table}[p]
    \centering
    \caption{PIQUE scores from CT CLADE (with $\mathcal{L}_{gmap}$) hyperparameter optimisation. The final chosen loss weighting is outlined in bold, based on PIQUE as calculated across all 15 volumes in the CT test dataset, shown in ascending PIQUE order, as well as visual image quality (see Figure \ref{fig:supporting_inf_4}). Each model is trained for 1 epoch, and images are assessed with a stride length of 6. The loss weightings are chosen from the following parameter sets: $\lambda_{gmap} \in \{0.1, 1, 10\}$, $\lambda_{cyc} \in \{0, 1, 10\}$ and $\lambda_{ident} \in \{1, 5, 10\}$. \\
   }
    \medskip
       \begin{tabular}{cccccc}
        \toprule
        $\lambda_{cyc}$ & $\lambda_{ident}$ & $\lambda_{gmap}$ & PIQUE ($\mu \pm \sigma$) \\
        \midrule
        0.1 & 0 & 1 & $27.65 \pm 4.45$ \\
        0.1 & 0 & 5 & $32.00 \pm 3.23$ \\
        0.1 & 0 & 10 & $34.78 \pm 3.66$ \\
        0.1 & 1 & 1 & $35.15 \pm 5.08$ \\
        0.1 & 1 & 5 & $31.23 \pm 3.06$ \\
        0.1 & 1 & 10 & $24.47 \pm 3.44$ \\
        0.1 & 10 & 1 & $35.72 \pm 4.20$ \\
        0.1 & 10 & 5 & $33.08 \pm 4.23$ \\
        0.1 & 10 & 10 & $30.73 \pm 3.80$ \\
        1 & 0 & 1 & $36.29 \pm 3.04$ \\
        1 & 0 & 5 & $28.26 \pm 4.92$ \\
        1 & 0 & 10 & $34.34 \pm 3.75$ \\
        1 & 1 & 1 & $33.67 \pm 5.86$ \\
        1 & 1 & 5 & $28.97 \pm 3.89$ \\
        \textbf{1} & \textbf{1} & \textbf{10} & $\mathbf{31.24 \pm 3.93}$ \\
        1 & 10 & 1 & $31.77 \pm 3.80$ \\
        1 & 10 & 5 & $33.51 \pm 4.42$ \\
        1 & 10 & 10 & $38.91 \pm 4.43$ \\
        10 & 0 & 1 & $30.86 \pm 4.18$ \\
        10 & 0 & 5 & $32.14 \pm 5.20$ \\
        10 & 0 & 10 & $36.45 \pm 5.27$ \\
        10 & 1 & 1 & $33.79 \pm 3.92$ \\
        10 & 1 & 5 & $36.91 \pm 5.27$ \\
        10 & 1 & 10 & $38.68 \pm 4.88$ \\
        10 & 10 & 1 & $37.91 \pm 4.87$ \\
        10 & 10 & 5 & $38.23 \pm 5.54$ \\
        10 & 10 & 10 & $42.68 \pm 5.46$ \\
        \bottomrule
    \end{tabular}
    \label{tab:supporting_inf_2}
\end{table}

\clearpage

\begin{table}[p]
    \centering
    \caption{PIQUE scores ($\mu \pm \sigma$) as calculated across all 15 volumes in the MRI test dataset, assessed with a stride length of 6. The scores are compared across different epochs for the final MRI CLADE (with $\mathcal{L}_{gmap}$) model ($\lambda_{cyc} = 1$, $\lambda_{ident} = 1$, $\lambda_{cyc} = 5$). The chosen number of epochs is outlined in bold, based on PIQUE score and visual image quality (see Figure \ref{fig:supporting_inf_7}).}
    \medskip
    \begin{tabular}{cc}
        \toprule
        Epoch & PIQUE ($\mu \pm \sigma$) \\
        \midrule
        1 & $33.73 \pm 3.57$ \\
        2 & $44.12 \pm 0.71$ \\
        3 & $23.34 \pm 4.75$ \\
        4 & $35.53 \pm 5.93$ \\
        5 & $26.52 \pm 4.68$ \\
        6 & $41.62 \pm 5.18$ \\
        7 & $39.08 \pm 5.05$ \\
        8 & $31.46 \pm 5.43$ \\
        \textbf{9} & $\mathbf{27.42 \pm 4.19}$ \\
        10 & $30.44 \pm 4.70$ \\
        \bottomrule
    \end{tabular}
    \label{tab:supporting_inf_5}
\end{table}

\begin{table}[p]
    \centering
    \caption{PIQUE scores ($\mu \pm \sigma$) as calculated across all 15 volumes in the CT test dataset, assessed with a stride length of 6. The scores are compared across different epochs for the final CT CLADE (with $\mathcal{L}_{gmap}$) model ($\lambda_{cyc} = 1$, $\lambda_{ident} = 1$, $\lambda_{cyc} = 10$). The chosen number of epochs is outlined in bold, based on PIQUE score and visual image quality (see Figure \ref{fig:supporting_inf_8}).}
    \medskip
    \begin{tabular}{cc}
        \toprule
        Epoch & PIQUE ($\mu \pm \sigma$) \\
        \midrule
        1 & $30.06 \pm 4.68$ \\
        2 & $36.28 \pm 4.33$ \\
        3 & $33.26 \pm 5.87$ \\
        4 & $37.76 \pm 3.90$ \\
        5 & $33.00 \pm 5.20$ \\
        6 & $34.23 \pm 5.13$ \\
        7 & $29.73 \pm 4.83$ \\
        8 & $37.67 \pm 3.99$ \\
        9 & $29.08 \pm 4.25$ \\
        \textbf{10} & $\mathbf{28.24 \pm 4.45}$ \\
        \bottomrule
    \end{tabular}
    \label{tab:supporting_inf_6}
\end{table}

\clearpage

\clearpage

\begin{table}[p]
    \centering
    \caption{Impact of stride length on image quality and inference time as assessed by PIQUE scores ($\mu \pm \sigma$) as calculated across all 15 volumes in the MRI test dataset. The scores are compared across different stride for the final MRI CLADE (with $\mathcal{L}_{gmap}$) model ($\lambda_{cyc} = 1$, $\lambda_{ident} = 1$, $\lambda_{cyc} = 5$, with 9 epochs). The chosen stride length is outlined in bold, based on PIQUE score, inference time and visual image quality (see Figure \ref{fig:supporting_inf_11}).}
    \medskip
    \begin{tabular}{ccc}
        \toprule
        Stride Length & PIQUE ($\mu \pm \sigma$) & Time for Inference per 3D Volume (s) ($\mu \pm \sigma$) \\
        \midrule
        6 & $27.42 \pm 4.19$ & $317 \pm 132$ \\
        8 & $29.17 \pm 4.09$ & $195 \pm 80$ \\
        \textbf{12} & $\mathbf{26.61 \pm 3.78}$ & $\mathbf{103 \pm 38}$ \\
        16 & $31.25 \pm 5.32$ & $72 \pm 25$ \\
        20 & $25.41 \pm 3.88$ & $53 \pm 16$ \\
        \bottomrule
    \end{tabular}
    \label{tab:supporting_inf_9}
\end{table}

\begin{table}[p]
    \centering
    \caption{Impact of stride length on image quality and inference time as assessed by PIQUE scores ($\mu \pm \sigma$) as calculated across all 15 volumes in the CT test dataset. The scores are compared across different stride for the final CT CLADE (with $\mathcal{L}_{gmap}$) model ($\lambda_{cyc} = 1$, $\lambda_{ident} = 1$, $\lambda_{cyc} = 10$, with 10 epochs). The chosen stride length is outlined in bold, based on PIQUE score, inference time and visual image quality (see Figure \ref{fig:supporting_inf_12}).}
    \medskip
    \begin{tabular}{ccc}
        \toprule
        Stride Length & PIQUE ($\mu \pm \sigma$) & Time for Inference per 3D Volume (s) ($\mu \pm \sigma$) \\
        \midrule
        \textbf{6} & $\mathbf{28.31 \pm 4.42}$ & $\mathbf{292 \pm 133}$ \\
        8 & $28.20 \pm 4.53$ & $183 \pm 74$ \\
        12 & $24.69 \pm 4.53$ & $100 \pm 34$ \\
        16 & $28.59 \pm 4.81$ & $73 \pm 19$ \\
        20 & $21.49 \pm 4.84$ & $54 \pm 16$ \\
        \bottomrule
    \end{tabular}
    \label{tab:supporting_inf_10}
\end{table}

\clearpage

\clearpage

{\small
\bibliographystyle{ieee_fullname}
\bibliography{main}
}

\end{document}